\documentclass{article}

\PassOptionsToPackage{numbers,sort&compress}{natbib}

\usepackage[final]{neurips_2021}

\usepackage[utf8]{inputenc} 
\usepackage[T1]{fontenc}    
\usepackage[colorlinks=true,linkcolor=black,citecolor=blue,urlcolor=blue,]{hyperref}
\usepackage{url}            
\usepackage{booktabs}       
\usepackage{amsfonts}       
\usepackage{nicefrac}       
\usepackage{microtype}      
\usepackage{xcolor}         

\usepackage{tablefootnote}
\usepackage[flushleft]{threeparttable}
\usepackage{wrapfig}
\usepackage{graphicx}
\usepackage{caption}
\usepackage{subcaption}
\usepackage{mathrsfs}
\usepackage{makecell}
\usepackage{etoolbox}
\usepackage{multirow}
\usepackage{stfloats}
\usepackage{amssymb}
\usepackage{amsmath}
\usepackage{mathtools}
\usepackage{array}
\usepackage{algorithm}
\usepackage{algorithmic}

\usepackage{bm}
\newcommand*{\B}[1]{\ifmmode\bm{#1}\else\textbf{#1}\fi}

\title{Learning to Iteratively Solve Routing Problems with Dual-Aspect Collaborative Transformer}

\author{%
Yining~Ma$^1$,%
~Jingwen~Li$^1$,%
~Zhiguang~Cao$^{2\text{,}}$\thanks{Zhiguang Cao and Wen Song are the corresponding authors.}~,%
~Wen~Song$^{3\text{,}*}$,%
~Le~Zhang$^4$,\\%
~\textbf{Zhenghua~Chen$^5$,}%
~\textbf{Jing~Tang}$^6$\\\\
$^{1}$National University of Singapore\\
$^{2}$Singapore Institute of Manufacturing Technology, A*STAR\\
$^{3}$Institute of Marine Science and Technology, Shandong University\\
$^{4}$University of Electronic Science and Technology of China\\
$^{5}$Institute for Infocomm Research, A*STAR\\
$^{6}$The Hong Kong University of Science and Technology\\
\texttt{\{yiningma, lijingwen\}@u.nus.edu, zhiguangcao@outlook.com,}\\
\texttt{wensong@email.sdu.edu.cn, zhangleuestc@gmail.com,}\\
\texttt{chen0832@e.ntu.edu.sg, jingtang@ust.hk}
}

\begin{document}

\maketitle

\begin{abstract}

Recently, Transformer has become a prevailing deep architecture for solving vehicle routing problems (VRPs). However, it is less effective in learning \emph{improvement} models for VRP because its positional encoding (PE) method is not suitable in representing VRP solutions. This paper presents a novel \textit{Dual-Aspect Collaborative Transformer} (DACT) to learn embeddings for the node and positional features separately, instead of fusing them together as done in existing ones, so as to avoid potential noises and incompatible correlations. Moreover, the positional features are embedded through a novel \textit{cyclic positional encoding} (CPE) method to allow Transformer to effectively capture the circularity and symmetry of VRP solutions (i.e., cyclic sequences). We train DACT using Proximal Policy Optimization and design a curriculum learning strategy for better sample efficiency. We apply DACT to solve the traveling salesman problem (TSP) and capacitated vehicle routing problem (CVRP). Results show that our DACT outperforms existing Transformer based improvement models, and exhibits much better generalization performance across different problem sizes on synthetic and benchmark instances, respectively.
\end{abstract}

\section{Introduction}
\label{sec:intro}
Vehicle Routing problems (VRPs), such as the Traveling Salesman Problem (TSP) and the Capacitated Vehicle Routing Problem (CVRP) which consider finding the optimal route for a single or fleet of vehicles to serve a set of customers, have ubiquitous real-world applications~\cite{toth2014vehicle, schneider2014electric}. Despite being intensively studied in the Operations Research (OR) community, VRPs still remain challenging due to their NP-hard nature~\cite{lenstra1981complexity}. Recent studies on learning neural heuristics are gathering attention as promising extensions to traditional hand-crafted ones (e.g., \cite{chen2019learning,kool2018attention,joshi2019efficient,xin2020multi,kwon2020pomo,xin2020step,zhang2020learning,wu2021learning,lu2019learning,hottunglearning,li2021deep}), where reinforcement learning (RL)~\cite{sutton2018reinforcement} is usually exploited to train a deep neural network as an efficient solver without hand-crafted rules. A salient motivation is that deep neural networks may learn better heuristics by identifying useful patterns in an end-to-end and data-driven fashion.

Solutions to VRPs, i.e., routes, are sequences of nodes (customer and depot locations). Naturally, deep models for Natural Language Processing (NLP), which deal with sequence data as well, are ideal choices for encoding VRP solutions. Given its remarkable performance in NLP tasks, Transformer \cite{vaswani2017attention} is standing at the forefront in the learning based methods for VRPs (e.g., \cite{kool2018attention,xin2020multi,kwon2020pomo,wu2021learning,lu2019learning,hottunglearning,li2021heterogeneous}). The original Transformer encodes a sentence, i.e., a sequence of words, into a unified set of embeddings by injecting word positional information into its word embeddings through positional encoding (PE). When it comes to VRPs, while is not required in \emph{construction} models, positional information is critical for deep models that learn \emph{improvement} heuristics since the input are solutions to be improved.

\begin{figure}
\centering 
     \subfloat[]{\includegraphics[width = 1.65in]{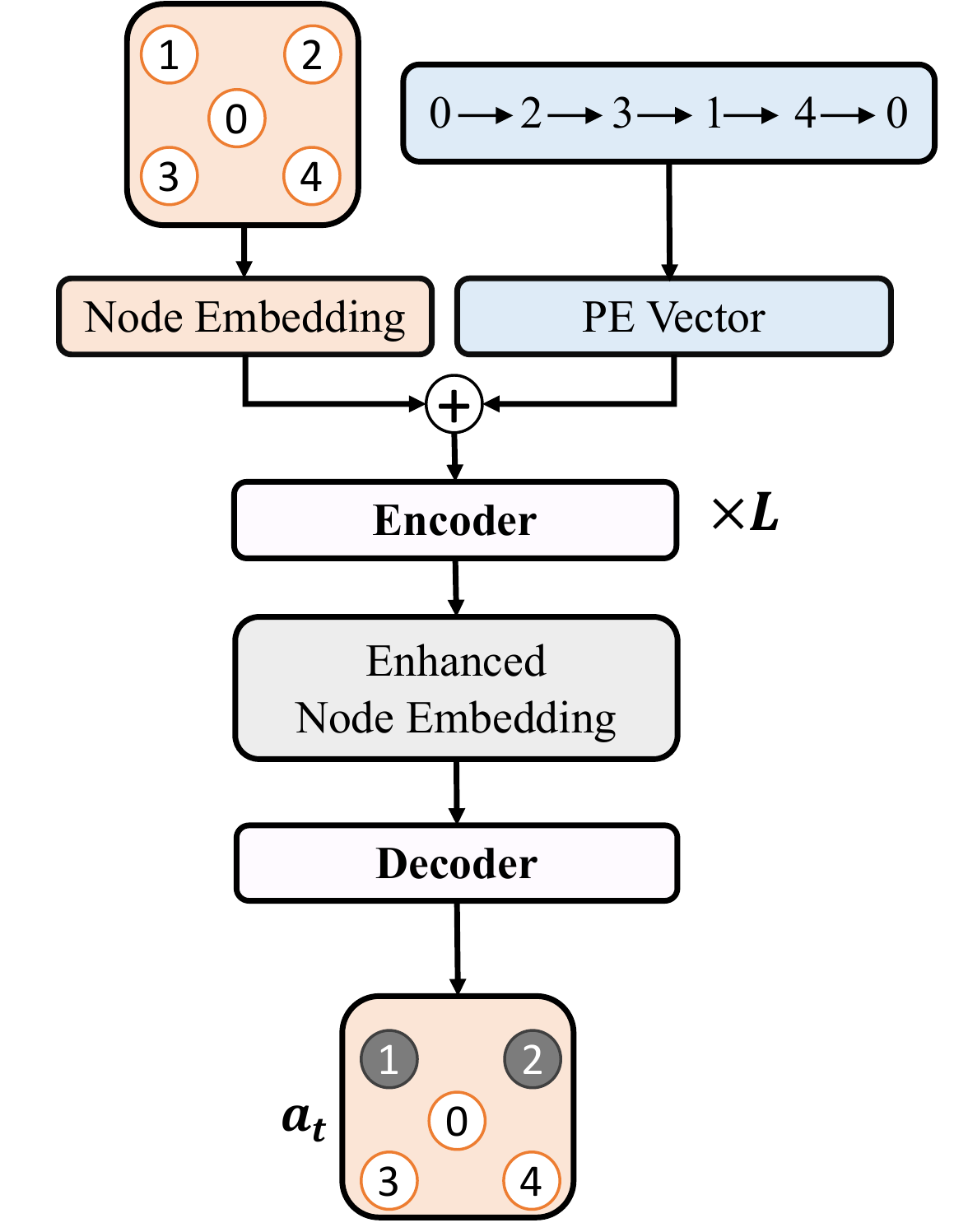}
     }
          \subfloat[]{\includegraphics[width = 1.8in]{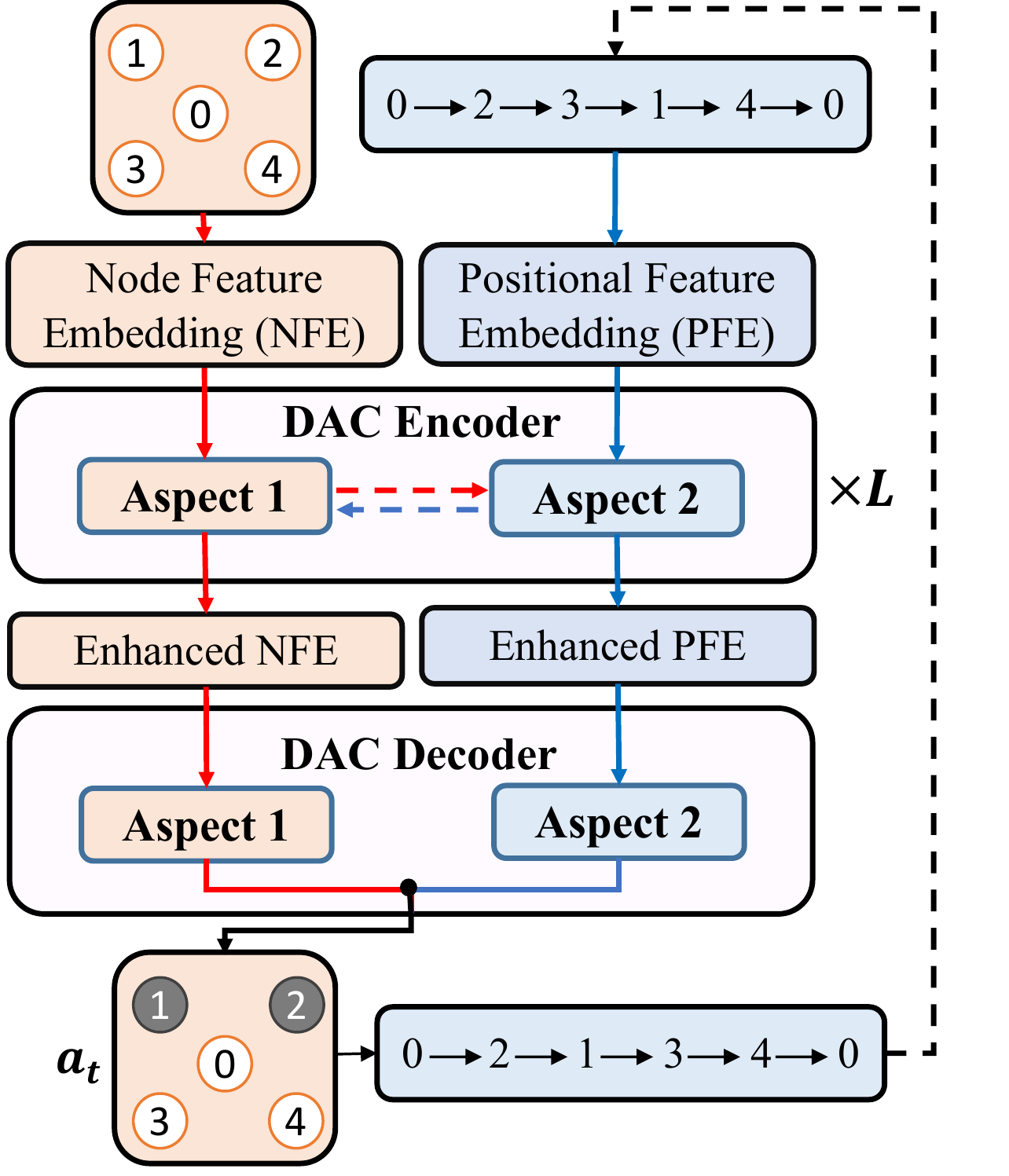}
     }
    \vspace{1pt}
    \caption{Transformer frameworks for VRPs. (a) \citet{wu2021learning} (the original one); (b) DACT (ours).} 
	\label{fig:framework} 
\end{figure}

Although some success has been achieved, learning \emph{improvement} heuristics for VRPs based on the original Transformer encoder is yet lacking from our perspective. Firstly, directly applying addition operation on PE vectors and the embeddings in absolute PE method (i.e., Figure \ref{fig:framework}(a)) could limit the representation of the model~\cite{ke2020rethinking}, as the mixed correlations\footnote{The term \emph{correlation} refers to the dot product between Query and Key in the self-attention module. The term \emph{mixed correlation} refers to the case where Query and Key are projected from different types of embeddings.} existing in the self-attention can bring unreasonable noises and random biases to the encoder (details in Appendix \ref{app:unified}).
Secondly, existing PE methods tend to fuse the node and positional information into one unified representation. 
NLP tasks such as translation may benefit from this owing to the deterministic and instructive nature of the positional information. However, such design may not be optimal for routing tasks because the positional information therein can be non-deterministic and sometimes even random. This may cause disharmony or disturbance in the encoder and may thus deteriorate the performance. 
Finally, most VRPs seek the shortest loop of the nodes, making their solutions to be cyclic sequences. However, existing PE methods are only designated to encode linear sequences\footnote{The relative PE method seems to help, however, it is found to be even worse than the absolute PE method for VRPs in~\citet{wu2021learning}, partly due to the disharmony issue caused by learning the unified representation.}, which may fail to identify such circular input. As will be shown in our experiments, this could severely damage the generalization performance, since the cyclic feature of VRP solutions is not correctly reflected by the encoder. 

In this paper, we address the above issues and contribute to the line of using RL to learn neural improvement heuristics for VRPs. We introduce the \textit{Dual-Aspect Collaborative Transformer (DACT)}, where we revisit the solution representations and propose to learn separated groups of embeddings for the node and positional features of a  VRP solution as shown in Figure \ref{fig:framework}(b). Our DACT follows the encoder-decoder structure. In the encoder, each set of embeddings encodes the solution mainly from its own aspect, and at the same time exploits a \textit{cross-aspect referential attention mechanism} for better perceiving the consistence and differentiation with respect to the other aspect. The decoder then collects action distribution proposals from the two aspects and synthesizes them to output the final one. Meanwhile, we design a novel \emph{cyclic positional encoding} (CPE) method to capture the circularity and symmetry of VRP solutions, which allows Transformer to encode cyclic inputs, and also boost the generalization performance for solving VRPs. As the last contribution,
we design a simple yet effective curriculum learning strategy to improve the sample efficiency. This further leads to faster and more stable convergence of RL training. Extensive experiments show that our DACT can outperform existing Transformer based improvement models with fewer parameters, and also generalizes well across different sizes of synthetic and benchmark instances, respectively.

\section{Related work}
\label{sec:relatedwork}
\subsection{Positional encoding (PE) in Transformer.}
The original Transformer adopted the absolute PE method to describe the absolute position of elements in the sequence \cite{vaswani2017attention}, especially for NLP. As formulated in Eq.~(\ref{eq:ape}), each generated positional embedding $p_i\!\in\!\mathbb{R}^{d}$ is added together with the $i$-th \emph{word} embedding $x_i$ in the first layer of the encoder,
\begin{equation}
\label{eq:ape}
    \alpha_{i,j}^{\text{Abs}} = \frac{1}{\sqrt{d}}((x_i + p_i)W^Q) ((x_j + p_j)W^K)^T.
\end{equation}
The relative PE method was further proposed in~\citet{shaw2018self} to better capture the relative order information. On the basis of absolute PE, it introduces an inductive bias to the attention as follows,
\begin{equation}
    \alpha_{i,j}^{\text{Rel}} = \frac{1}{\sqrt{d}}((x_i + p_i)W^Q) ((x_j + p_j)W^K + a_{j-i})^T,
\end{equation}
where $a_{j-i}\!\in\!\mathbb{R}^{d}$ is learnable parameters for encoding the relative position $j\!-\!i$.
To avoid the mixed and noisy correlations between word semantics and positional information in the above two PEs, the Transformer with United Positional Encoding (TUPE) \cite{ke2020rethinking} was proposed for NLP which utilizes separated projection metrics $W_x$ and $W_p$ for each information as follows,
\begin{equation}
    \alpha_{i,j}^{\text{TUPE}} = \frac{1}{\sqrt{2d}}(x_iW_x^Q) (x_jW_x^K)^T + \frac{1}{\sqrt{2d}}(p_iW_p^Q) (p_jW_p^K)^T + b_{j-i}.
\end{equation}
However, as mentioned previously, existing PE methods are less effective for VRPs since they simply fuse the node and positional information into one unified set of embeddings during or after the calculation of the attention correlation $\alpha_{i,j}$. Meanwhile, they are also unable to properly encode and handle cyclic input sequences as in VRP solutions.

\subsection{Deep models for VRP.} Various deep architectures such as Recurrent Neural Network (RNN), Graph Neural Network (GNN), and Transformer have been employed in solving VRPs.

\textbf{RNN based models.} As the pioneering work of neural VRP solvers, Pointer Network adopted RNN and supervised learning to solve TSP~\cite{vinyals2015pointer} (extended to RL in~\citet{bello2016neural} and CVRP in~\citet{nazari2018reinforcement}). While the models in \cite{vinyals2015pointer, d2020learning,bello2016neural,nazari2018reinforcement} learn \textit{construction} heuristics, NeuRewriter \cite{chen2019learning} learns \textit{improvement} heuristic for CVRP using LSTM to encode the positional information of a solution. In~\citet{hottunglearning}, the conditional variational autoencoder was adopted to learn a continuous and latent search space for VRP, where high-quality solutions were taken as input and encoded by RNNs. However, recurrence structures in RNN are less efficient in both representation and computation \cite{kool2018attention}.

\textbf{GNN based models.} 
In \citet{dai2017learning}, GNN was combined with Q-learning for solving TSP. Based on supervised learning, \citet{joshi2019efficient} 
used GNN to learn heatmaps that prescribe the probability of each edge appearing in the optimal TSP tour.
This idea was extended in \citet{fu2020generalize} with additional components such as graph sampling and heatmap merging to enable generalization to larger TSP instances. These models often require post-processing to construct feasible solutions from heatmaps (e.g., beam search \cite{joshi2019efficient}, Monte-Carlo tree search \cite{fu2020generalize}, and dynamic programming \cite{kool2021deep}).

\textbf{Transformer based models.} 
The Attention Model (AM) by~\citet{kool2018attention} was recognized as the first success of Transformer based models for VRPs. Based on AM, \citet{xin2020multi} proposed a Multi-Decoder AM that learns multiple diverse policies for better performance. In \citet{kwon2020pomo}, the RL algorithm of AM was improved which leaded to a new solver, i.e., POMO (Policy Optimization with Multiple Optima), and achieved the state-of-the-art performance. However, POMO is still lacking in generalization. Besides these \textit{construction} models, Transformer was also explored to learn \textit{improvement} heuristics. \citet{hottung2019neural} learned first neural large neighborhood search algorithm for VRPs. \citet{lu2019learning} proposed the L2I model that learns to select local search operators from a pool of traditional ones. Both methods used a Transformer-style encoder, but the positional information is captured in the node features (information of previous and next nodes) instead of using PE methods. Though L2I was shown to outperform LKH3~\cite{lkh3}, it is limited to CVRP and the required time is considerably long. \citet{wu2021learning} proposed a Transformer model which learns to pick node pair in each step to perform a pairwise local operator (e.g., 2-opt). However, it suffers from the inaccurate representation of positional information given the original Transformer encoder.

\section{Problem formulation}
\label{sec:model}
We define a VRP instance as a group of $N$ nodes to visit, where the \textit{node feature} $x_i$ of node $i$ contains 2-dim coordinates and other problem-specific features (e.g., customer demand). A solution $\delta$ consists of a sequence of nodes visited in order where we denote $p_i$ to be the position (indices) of node $i$ in the solution which is deemed as the \textit{positional feature} of node $i$. The objective is to minimize the total travel distance $D(\delta)$ under certain problem-specific constraints.

\begin{wrapfigure}{r}{2.6in}
\centering 
    \vspace{-5pt}
	\includegraphics[width=2.5in]{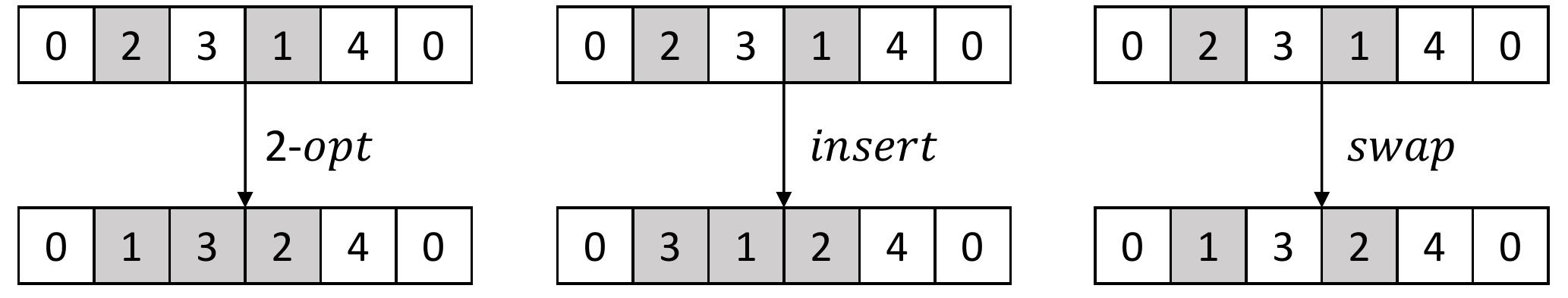}
	\vspace{5pt}
	\caption{Illustration examples of three pairwise operators for routing problems when node pair ($i=2$, $j=1$) is specified for operating. From left to right: \emph{2-opt}, \emph{insert}, and \emph{swap}.} 
	\label{fig:operation}
\end{wrapfigure}

Starting with an initial yet complete solution, our neural RL policy tries to improve the solution iteratively. At each step, the policy automatically selects a pair of nodes and locally adjusts the solution using a preset pairwise operator such as \emph{2-opt}, \emph{insert}, or \emph{swap}. As illustrated in Figure~\ref{fig:operation}, given a node pair ($i,j$), the \emph{2-opt} operator adjusts a solution by reversing the segment between node $i$ and node $j$; the \emph{insert} operator adjusts a solution by placing node $i$ after node $j$; and the \emph{swap} operator adjusts a solution by exchanging the position of node $i$ and node $j$. Such operation is repeated until reaching the step limit $T$ and we model it in the form of Markov Decision Process (MDP) as follows.

\begin{figure*}
\centering 
	\includegraphics[width=0.99\textwidth]{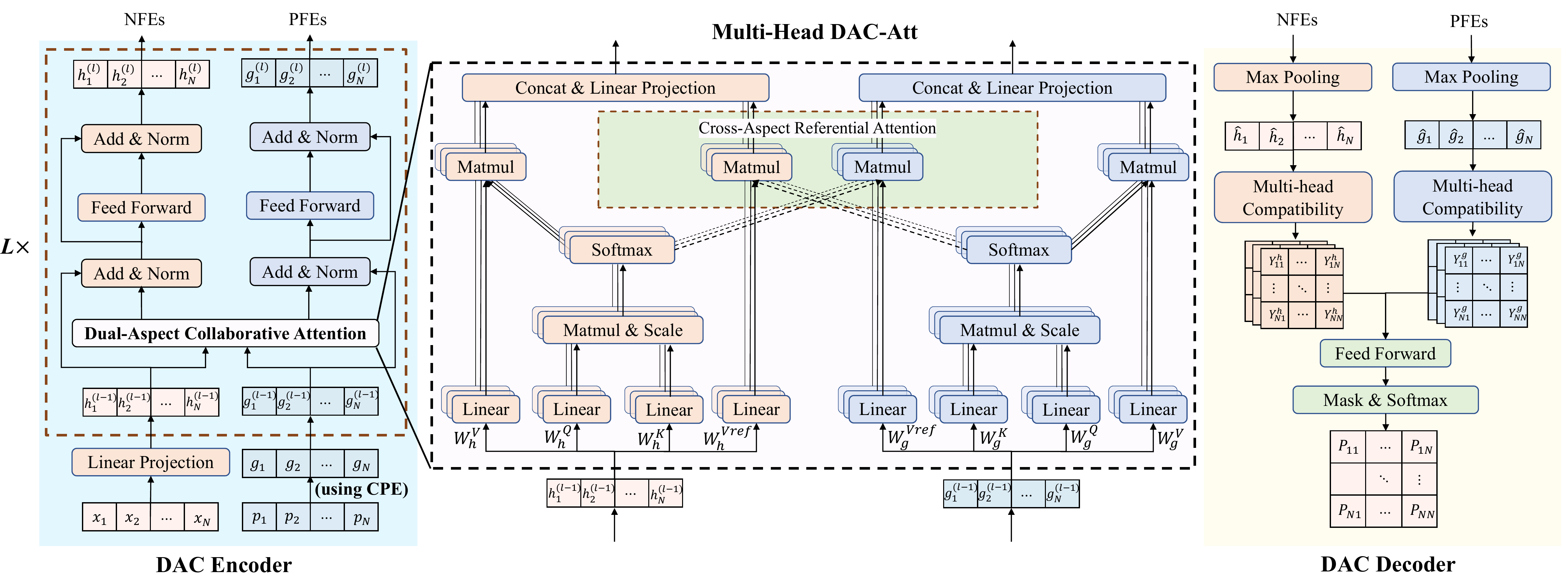}
	\vspace{5pt}
	\caption{Architecture of our policy network,  dual-aspect collaborative Transformer (DACT).}
	\label{fig:DL_model}
\end{figure*}

\textbf{State.} 
For an instance with $N$ nodes, 
a state describes current solution $\delta_t$ using its node and positional features of each node, i.e., $s_t = \Psi(\delta_{t}) = \{x_{1}^t, ...,  x_{N}^t, p_{1}^t,..., p_{N}^t\}$.

\textbf{Action.}
The action $a_t = (i,j)$ specifies  a node pair $(i,\!j)$ for the pairwise operator.

\textbf{Reward.} 
The reward function is defined as, $r_t\!=\!D(\delta_t^{\textit{*}})\!-\!min\left[D(\delta_{t+1}), D(\delta_t^{\textit{*}})\right]$ where $\delta_t^{\textit{*}}$ is the best incumbent solution found until time $t$. It refers to the immediate reduced cost at each step with respects to the best incumbent solution, which ensures the cumulative reward equal to the total reduced cost over the initial solution. Hence the reward $r_t\!>\!0$ if and only if a better solution is found.

\textbf{Policy.}
The policy $\pi_\theta$ is parameterized by the proposed DACT model with parameters $\theta$. At each time step, the action $(i,j)$ is obtained by sampling the stochastic policy for both training and inference.

\textbf{Transition.} 
The next state $s_{t+1}$ is originated from $s_t$ by performing the preset pairwise operator on the given node pair (action). Our state transient is deterministic, in the sense that it always accepts the next solution as the next state (infeasible solutions will be masked), regardless of its objective value. With such simple rule, the RL agent is expected to automatically learn how to combine multiple steps of simple local movements to achieve better solutions, even if some of them may worsen the current solution. 
Note that the step limit $T$ can be any user-specified value according to the allowed time budget. Hence, our MDP can have infinite horizon and we consider the reward discount factor $\gamma\!<\!1$. Besides, we stipulate that if $T_r$ (a hyper-parameter) subsequent steps do not yield a better solution, we will reinitialize the search by assigning the best solution found thus far as the next solution to move on.

\section{Dual-aspect collaborative Transformer model}
\label{sec:methodology}
We now present the details of our \textit{Dual-Aspect Collaborative Transformer} (DACT). The concrete architecture of DACT is presented in Figure \ref{fig:DL_model}, where we take the TSP with $N$ nodes as an illustration example. Our DACT leverages separate \emph{aspects} of embeddings to encode a VRP solution. In the DAC \emph{encoder}, the self-attention correlations are computed individually for each aspect, and a \emph{cross-aspect referential attention mechanism} is proposed to enable one aspect to effectively exploit attention correlations from the other aspect as optional references. The DAC \emph{decoder} then collects action distribution proposals from both aspects and synthesize them to the final one.

\subsection{Dual-aspect solution representation}

Specifically, we propose to learn two sets of embeddings, i.e., the \textit{node feature embeddings} (NFEs) for node representation and the \textit{positional feature embeddings} (PFEs) for positional representation.

\begin{wrapfigure}{r}{2.6in}
\centering 
    \vspace{-15pt}
	\includegraphics[width=2.6in]{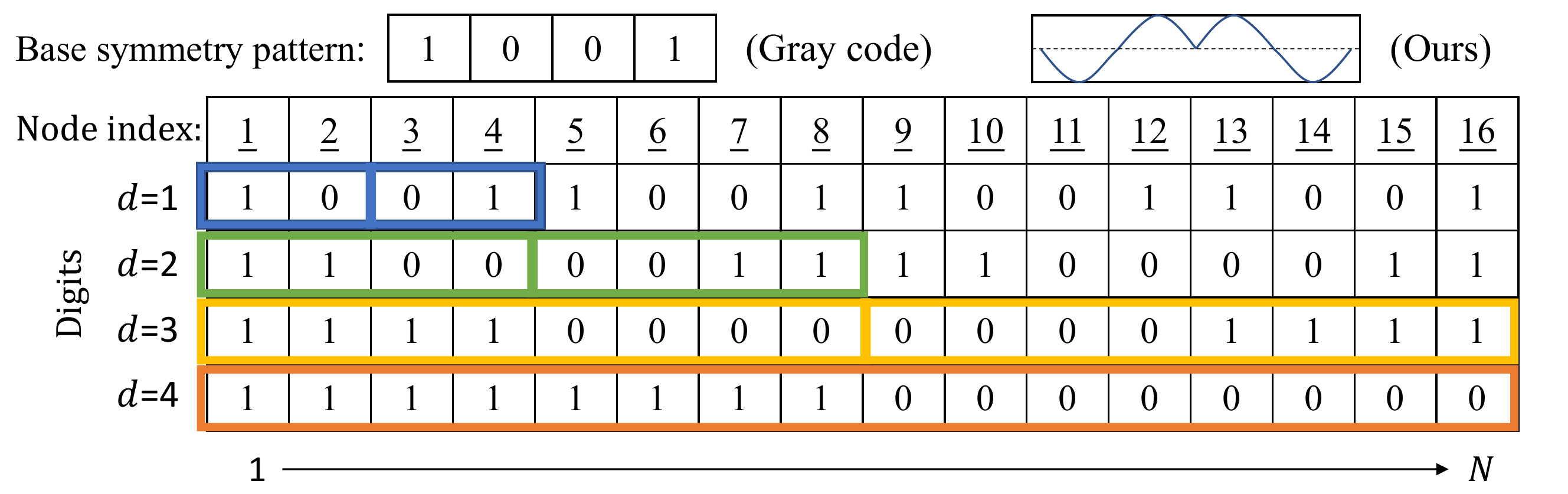}
	\caption{An example of cyclic Gray code where 4 digits are used to encode $N\!=\!16$ nodes. The top left shows the base symmetry pattern `1001' in Gray code, and the top right plots its representation in our method.} 
	\label{fig:Graycode}
	\vspace{-15pt}
\end{wrapfigure}

\textbf{NFEs.} 
Following~\cite{kool2018attention,wu2021learning}, the NFE $h_i$ of node $i$ is initialized as the linear projection of its node feature $x_i$ with output dimension\footnote{Different from~\citet{wu2021learning}, we reduce the dimension of the embeddings from 128 to 64.} $dim=64$.

\textbf{PFEs.}
The PFE $g_i$ of the positional feature $p_i$ is initialized as a real-valued vector ($dim =64$) by applying our cyclic positional encoding (CPE), which is designed based on cyclic Gray codes \cite{wiki:Gray_code}.

\begin{figure}
     \centering
     \subfloat[]{\includegraphics[width = 0.31\textwidth]{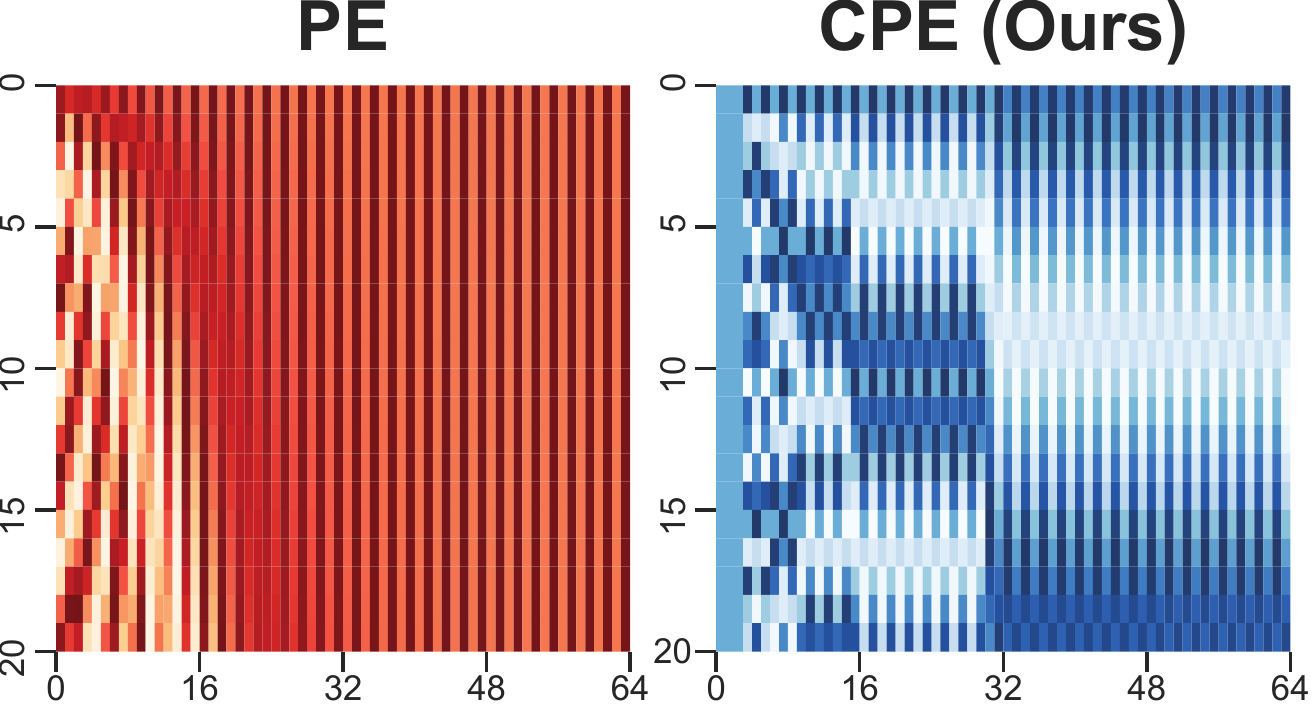}
     \label{<figure1>}}
     \hspace{5pt}
     \subfloat[]{\includegraphics[width = 0.31\textwidth]{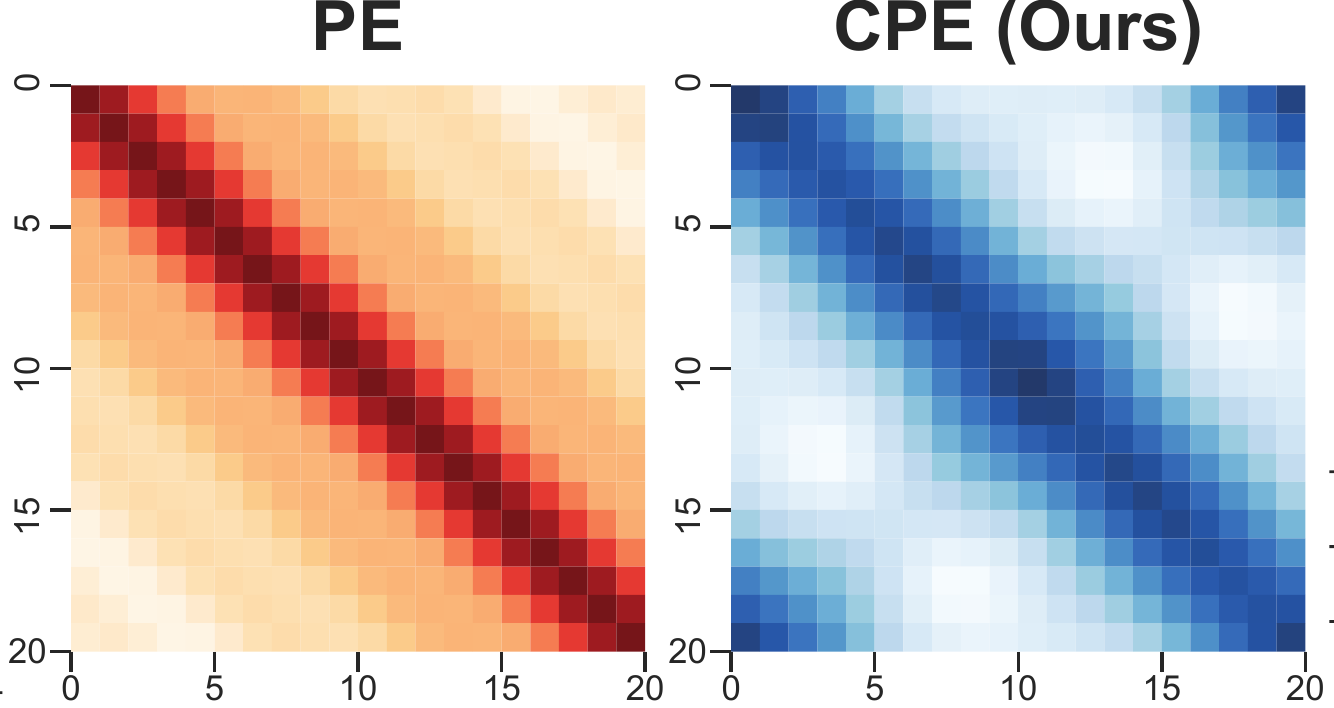}
     \label{<figure2>}}
     \hspace{5pt}
     \subfloat[]{\includegraphics[width = 0.31\textwidth]{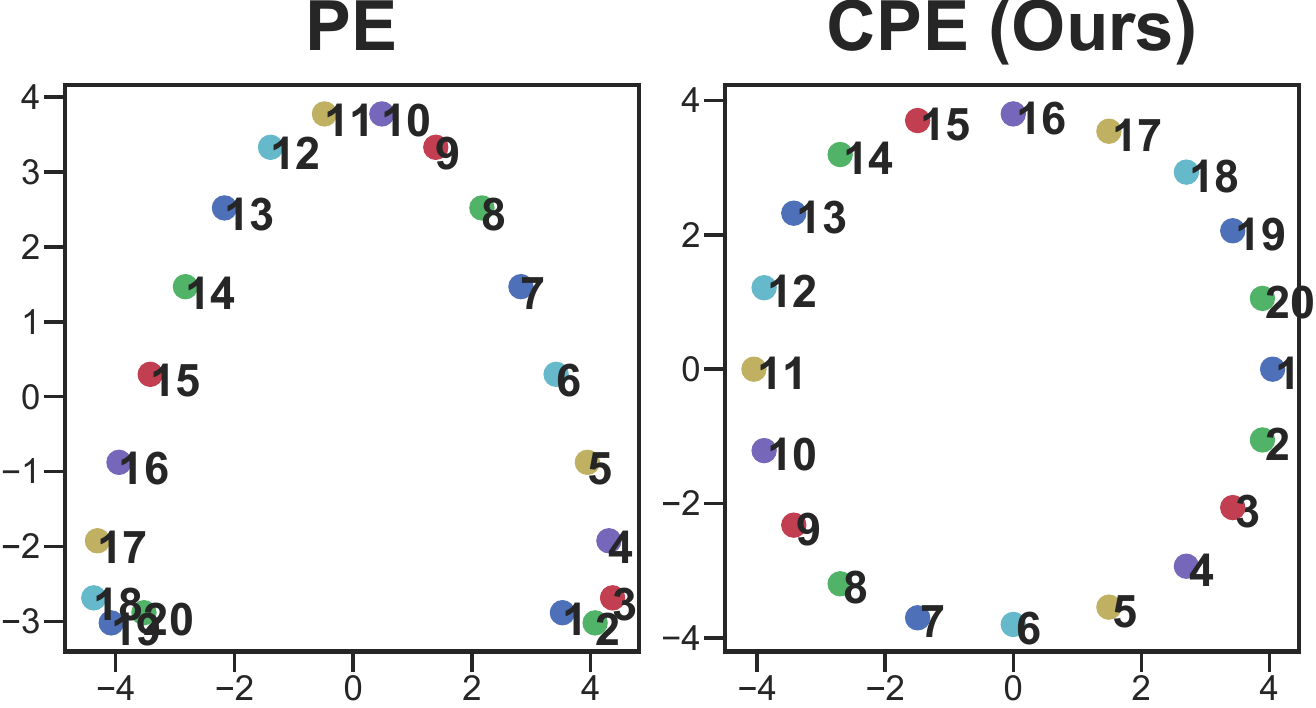}
     \label{<figure3>}}

     \caption{Comparison of our CPE method with absolute PE method on a TSP instance with 20 nodes. (a) the embedding vectors, (b) the correlations (dot products) between every two embeddings, and (c) the top two principal components after PCA (principal component analysis) projection.}
     \label{fig:pos_enc_pattern}
\end{figure}

As illustrated in Figure \ref{fig:Graycode}, the cyclic Gray codes present a cyclic property (`1110' in the last column is adjacent to `1111' in the first column) and an adjacency similarity property (any codes in adjacent columns only differ in one digit), both of which are desirable for cyclic sequences. To preserve these properties in designing our CPE, we follow two observed patterns: 1) each numerical digit contains a periodic cycle with reflectional symmetry, e.g., the `10$|$01' in the lowest digit; and 2) the higher the numerical digit, the longer the period. Accordingly, we create similar patterns based on the sinusoidal functions in Eq.~(\ref{eq:cpe}), where a periodic function with period $\frac{4\pi}{\omega_d}$ (induced by modulus) is used to generate one base symmetry pattern (the top right in Figure~\ref{fig:Graycode}),
\begin{equation}
\label{eq:cpe}
  \overrightarrow{g_i}^{(d)} \coloneqq \left\{\begin{matrix}
\!sin(\omega_d\!\cdot\!\left| (z(i) \bmod \frac{4\pi}{\omega_d}) - \frac{2\pi}{\omega_d} \right|), \text{ if }d\text{ is even}\\
\!cos(\omega_d\!\cdot\!\left| (z(i) \bmod \frac{4\pi}{\omega_d}) - \frac{2\pi}{\omega_d} \right|), \text{ if }d\text{ is odd} \end{matrix}\right.
\end{equation}
where $ z(i) = \frac{i}{N}\frac{2\pi}{\omega_d}\left \lceil \frac{N}{2\pi/\omega_d} \right \rceil$ is to make $N$ nodes linearly spaced in the generated pattern; the angular frequency $\omega_d$ is decreasing along the dimension to make the wavelength longer within the range $[N^{\frac{1}{\left \lfloor dim / 2 \right \rfloor}}, N]$ (see Appendix~\ref{app:CPE} for details). 
In Figure~\ref{fig:pos_enc_pattern}, we visualize the comparison between the absolute PE and our CPE for encoding a TSP instance of 20 nodes. Figure~\ref{fig:pos_enc_pattern}(a) demonstrates that our real-valued base symmetry pattern has a longer cyclic period as the digit grows. Figure~\ref{fig:pos_enc_pattern}(b) indicates that our method (blue) is able to correctly reflect the adjacency between the head and tail of the cyclic sequence whereas the PE method (red) fails to do so. Figure~\ref{fig:pos_enc_pattern}(c) verifies that our CPE vectors are well distributed in space with desired cyclic and adjacency similarity properties.

\subsection{The encoder}
The encoder consists of $L\!=\!3$ stacked DAC encoders. In each DAC encoder, we retain relatively independent encoding stream for NFEs and PFEs as in Eq.~(\ref{eq:pathh}) and Eq.~(\ref{eq:pathg}), respectively, each of which consists of a shared Dual-Aspect Collaborative Attention (\textbf{DAC-Att}) sub-layer and an independent feed-forward network (\textbf{FFN}) sub-layer. DAC-Att takes both sets of embeddings as input and then outputs their respective enhanced embeddings, i.e., NFEs $\{\B{\tilde h}\}_{i=1}^N$ and PFEs $\{\B{\tilde g}\}_{i=1}^N$. Each sub-layer is followed by skip connection \cite{he2016deep} and layer normalization \cite{DBLP:journals/corr/BaKH16} as same as the original Transformer.
\begin{equation}
\label{eq:pathh}
    h_i^{(l)}\!=\!\textbf{LN}\!\left( {h}'_i + \textbf{FFN}_h^{(l)}\!\left ( {h}'_i \right ) \right),
    {h}'_i\!=\!\textbf{LN}\!\left( h^{(l\!-\!1)}_i + {\B{\tilde h}}^{(l)}_i\right),
\end{equation}
\begin{equation}
\label{eq:pathg}
    g_i^{(l)}\!=\!\textbf{LN}\!\left( {g}'_i + \textbf{FFN}_g^{(l)}\!\left ( {g}'_i \right ) \right),
    {g}'_i\!=\!\textbf{LN}\!\left( g^{(l\!-\!1)}_i + {\B{\tilde g}}^{(l)}_i\right).
\end{equation}

\textbf{DAC-Att.}
The DAC-Att sub-layer enhances each set of embedding from its own aspect, while leveraging attention correlations from the other aspect to achieve the synergy. Given the two sets of embeddings\footnote{We omit the encoder index $l$ for better readability.}, $\{h_i\}_{i=1}^N$ and $\{g_i\}_{i=1}^N$, we first compute the self-attention correlation from both aspects,
\begin{equation}
\label{eq:score}
    \alpha_{i,j}^{h} = \frac{1}{\sqrt{d_k}} \left( h_i W^{Q}_h \right ) \left( h_j W^{K}_h \right ) ^T, \hspace{0.4cm}
    \alpha_{i,j}^{g} = \frac{1}{\sqrt{d_k}} \left( g_i W^{Q}_g \right ) \left( g_j W^{K}_g \right ) ^T,
\end{equation}
where independent matrices $W_h^Q, W_h^K, W_g^Q$ and $W_g^K \in \mathbb{R}^{dim \times d_k}$ are used to calculate \emph{queries} and \emph{keys}. The obtained correlations are further normalized to $ \tilde \alpha_{i,j}^{h}$ and $ \tilde \alpha_{i,j}^{g}$ via Softmax. Note that the correlations are computed from their own aspect, which eliminates possible noises and conduces to correctly describe the incompatible node pair relationships in different aspects of VRP solutions.

We then exploit a \textit{cross-aspect referential attention} mechanism, which  allows computed correlations to be shared between each other, as additional references for both contradistinction and collaboration,
\begin{equation}
    \text{out}^h_i\!=\!\text{Concat}\!\left[ \sum_{j=1}^N{\left.\!\tilde\alpha_{i,j}^{h}\left( h_j W^{V}_h \right)\right.}\!, 
    \sum_{j=1}^N{\left.\!\tilde\alpha_{i,j}^{g}\!\left( h_j W^{\textit{Vref}}_h\right)\!\right.}\! \right],
\end{equation}
\begin{equation}
    \text{out}^g_i\!=\!\text{Concat}\!\left[ \sum_{j=1}^N{\left.\!\tilde\alpha_{i,j}^{g}\left( g_j W^{V}_g \right)\right.}, 
    \sum_{j=1}^N{\left.\!\tilde\alpha_{i,j}^{h}\left( g_j W^{\textit{Vref}}_g\right)\right.}\! \right],
\end{equation}
where $W_h^V, W_g^V \in \mathbb{R}^{dim \times d_v}$ are trainable parameter matrices for formulating \emph{values} in each aspect; and $W_h^\textit{Vref}, W_g^\textit{Vref} \in \mathbb{R}^{dim \times d_v}$ are parameter matrices for each aspect to generate \emph{referential values}. We finally use the \emph{multi-head attention} to get NFEs ${\tilde h_i}$ and PFEs ${\tilde g_i}$ as follows,
\begin{equation}
\begin{split}
{\tilde h}_i, {\tilde g}_i = \textbf{DAC-Att}\left(W^Q, \right. & \left. W^K, W^V, W^{V_\textit{ref}}, W^O\right),\\
    {\tilde h}_i  =  \text{Concat} \left[ \text{head}^h_{i, 1}, ..., \text{head}^h_{i, m} \right] W^O_h, &\hspace{0.2cm}
    {\tilde g}_i  =  \text{Concat}\left[ \text{head}^g_{i, 1}, ..., \text{head}^g_{i, m} \right] W^O_g,
\end{split}
\end{equation}
where $\text{head}_{i, k}^h\!=\!out_{i,k}^h$, $\text{head}_{i, k}^g\!=\!out_{i,k}^g$, and $W_h^O, W_g^O \in \mathbb{R}^{2md_v \times dim}$ are trainable parameter matrices. In our model, we adopt $m=4$ and $d_k = d_v = 16$.

\textbf{FFN.} Our FFN sub-layer has only one hidden layer with 64 hidden unites and adopts the ReLU activation function. The parameters of $\textbf{FFN}_h$ and $\textbf{FFN}_g$ are different for each group of embeddings.

\subsection{The decoder}
In the DAC decoder, the two sets of embeddings $\{h^{(L)} _i\}_{i=1}^N$ and $\{g^{(L)}_i\}_{i=1}^N$ are first passed through a \textbf{Max-pooling} sub-layer and a multi-head compatibility \textbf{(MHC)} sub-layer to independently generate diversified node-pair selection proposals from their own aspect, which are then aggregated through a feed-forward aggregation (\textbf{FFA}) sub-layer for output.

\textbf{Max-pooling}.
For each set of embeddings, we adopt the max-pooling sub-layer in \citet{wu2021learning} to aggregate the global representation of all $N$ embeddings into each respective one\footnote{E.g., for $\hat h$, $
    {\hat h}_i\!=\!h^{(L)}_iW_h^{\textit{Local}}\!+\!\max\left[\! \{h_i^{(L)}\}_{i=1}^N\!\right]W_h^{\textit{Global}} $,
where $W_h^{\textit{Local}},\!W_h^{\textit{Global}}\!\in\! \mathbb{R}^{64 \times 64}$ are parameters.}.

\textbf{MHC}. The compatibility sub-layer computes the attention correlations for each embedding pair, where the obtained correlations with size $N\times N$ will be deemed as a proposal distribution for node pair selection. Our correlations are computed based on multiple heads for diversity. And we calculate separated attention score matrices $Y^{h}_k , Y^{g}_k \in \mathbb{R}^{N \times N}$ (of head $k$) from the two aspects independently. Accordingly, the action distribution proposals would be different due to their aspect-specific focus and cognitions of the current solution, which will provide the subsequent FFA layer with a rich pool of proposals and allow our model to be more flexible and robust.

\textbf{FFA}.
Once all proposals from two aspects are collected, a FFN with four layers (dimensions are $2m$, 32, 32 and 1, respectively) and ReLU activation is used to aggregate them,
\begin{equation}
 \tilde Y_{i,j} = \textbf{FFA}\left( Y^{g}_{i,j,1},..., Y^{g}_{i,j,m}, Y^{h}_{i,j,1},... Y^{h}_{i,j,m} \right),
\end{equation}
where $m\!=\!4$ is the number of heads; and the output ${\tilde Y}_{i,j}$ is a scalar indicating the likelihood of selecting node pair $(i,j)$ as an action. Afterwards, we apply ${\hat Y}_{ij} = C \cdot \text{Tanh}({\tilde Y}_{i,j})$ with $C=6$ to control the entropy, and mask \footnote{Besides, we also mask all the diagonal elements since they are not meaningful to the pair-wise operators, and the node pair selected at the last step to forbid possible dead loops~\cite{wu2021learning}.} the infeasible node pairs ($i',j'$) as ${\hat Y}_{i'\!j'}\!=\!-\infty$. Lastly, the likelihoods are normalized using Softmax function to obtain the final action distribution $P_{i,j}$.

\subsection{Reinforcement learning algorithm}
\label{ppo-cl}

We adopt the proximal policy optimization \cite{schulman2017proximal} with $n$-step return estimation for training (details are given in Appendix \ref{app:algorithm}), and design a \textit{curriculum learning} (CL) strategy for better sample efficiency.

\textbf{Curriculum learning strategy.}
 The strategy in~\citet{wu2021learning} sets a maximum of $T_{train}$ steps for training and estimates future returns by \textit{bootstrapping} \cite{pardo2018time}. However, due to the concern of training cost, $T_{train}$ is usually much smaller than actual $T$ for inference (e.g., 200 v.s. 10k), which may leave the agent a poor chance of observing high-quality solutions (states) during training. Consequently, it may cause high variance for bootstrapping because the value function is mostly fitted on low-quality solutions and may render it less knowledgeable in estimating long-term future returns accurately. In this paper, we tackle this issue by a simple yet efficient strategy which gradually prescribes higher-quality solutions as the initial states for training. In doing so, 1) it increases the probability for the agent to observe better solutions and thus reduce the variance of the value function; 2) it increases the difficulty of the learning task (higher-quality solutions are harder to improve) in a gradual manner and achieves better sample efficiency~\cite{bengio2009curriculum}. In practice, those higher-quality solutions can be easily achieved by improving the randomly generated ones using the current policy for a few $T_{init}$ steps, where $T_{init}$ could be slightly increased as the epoch grows.

\section{Experiments}
\label{sec:exp}
\begin{table}
\centering
\caption{Comparison with various baselines on TSP and CVRP.}
\vspace{6pt}
\label{tab:main}
\resizebox{0.99\textwidth}{!}{%
\begin{threeparttable}
\begin{tabular}{cl||ccl|ccl|ccl}
\toprule
\multicolumn{2}{c||}{\multirow{2}{*}{Method}} & \multicolumn{3}{c|}{N=20} & \multicolumn{3}{c|}{N=50} & \multicolumn{3}{c}{N=100} \\
\multicolumn{2}{c||}{} & Obj. & Gap & Time$^*$ & Obj. & Gap & Time$^*$ & Obj. & Gap & Time$^*$ \\ 
\midrule
\midrule
\multirow{19}{*}{\rotatebox{90}{TSP}} 
& {Concorde}~\cite{concorde} & 3.83  &  -  & -  & 5.70 & -  & -  & 7.76  & -  & - \\
& {LKH}~\cite{lkh2} & 3.83  &  0.00\%  & {(6m)}   & 5.70 &  0.00\%  & {(1.5h)}  & 7.76  & 0.00\%  & {(6h)} \\
& OR-Tools~\cite{ortools} & 3.86$^\ddagger$  &  0.94\%$^\ddagger$  & (1m)$^\ddagger$  & 5.85$^\ddagger$ &  2.87\%$^\ddagger$  & (5m)$^\ddagger$ & 8.06$^\ddagger$  & 3.86\%$^\ddagger$  & (23m)$^\ddagger$  \\
\cmidrule(lr){2-11}
& {Neural-2-Opt \cite{d2020learning} (T=3k)} & {3.83}  &  \textbf{{0.00\%}}  & {(47m)} & {5.70} &  {0.10\%}  & {(1h)}  & {7.81}  & {0.61\%}  & {(1.5h)}\\ 
& {Neural-2-Opt \cite{d2020learning} (T=6k)} & {3.83}  &  \textbf{{0.00\%}}  & {(1.5h)}  & {5.70} &  {0.06\%}  & {(2h)}  & {7.79}  & {0.39\%}  & {(3.5h)}\\ 
& {\citet{wu2021learning} (T=1k)} & 3.83$^\ddagger$  &  0.03\%$^\ddagger$  & (12m)$^\ddagger$  & {5.74} &  {0.82\%}  & {(16m)}  & {8.02}  & {3.31\%}  & {(24m)}\\ 
& {\citet{wu2021learning} (T=5k)} & 3.83$^\ddagger$  &  \textbf{0.00\%}$^\ddagger$  & (1h)$^\ddagger$  & {5.71} &  {0.24\%} & {(1.5h)}  & {7.88}  & {1.54\%}  & {(2h)}\\ 
& DACT (T=1k) & 3.83  &  0.04\%  & {(2m)$[$40s$]$}  & 5.70 &  0.14\%  & {(7m)$[$2m$]$}  & 7.89  & 1.62\%  & {(20m)$[$5m$]$}  \\
& DACT (T=5k) & 3.83  &  \textbf{0.00\%}  & {(10m)$[$3m$]$} & 5.70  &  0.02\%  & {(32m)$[$8m$]$}  & 7.81  & 0.61\%  & {(1.5h)$[$25m$]$}   \\
& DACT (T=10k) & 3.83  &  \textbf{0.00\%}  & {(21m)$[$7m$]$}   & 5.70  &  0.01\%  & {(1h)$[$17m$]$}  & 7.79  & 0.37\%  & {(3.5h)$[$50m$]$}  \\
& DACT$\times$4 augment & 3.83  &  \textbf{0.00\%}  & {(1.5h)$[$28m$]$}  & 5.70  &  \textbf{0.00\%}  & {(4h)$[$1h$]$}  & 7.77  & \textbf{0.09\%}  & {(14h)$[$3.5h$]$} \\
\cmidrule(lr){2-11}
& GCN-BS \cite{joshi2019efficient} & 3.84$^\ddagger$
&  0.01\%$^\ddagger$  & (12m)$^\ddagger$  & 5.70$^\ddagger$ &  0.01\%$^\ddagger$  & (18m)$^\ddagger$  & 7.87$^\ddagger$  & 1.39\%$^\ddagger$  & (40m)$^\ddagger$  \\
& AM-sampling  \cite{kool2018attention} & 3.84$^\ddagger$ &  0.08\%$^\ddagger$  & (5m)$^\ddagger$  & 5.73$^\ddagger$ &  0.52\%$^\ddagger$  & (24m)$^\ddagger$  & 7.94$^\ddagger$  & 2.26\%$^\ddagger$  & (1h)$^\ddagger$\\ 
& MDAM-BS  \cite{xin2020multi} & 3.84$^\ddagger$  &  \textbf{0.00\%}$^\ddagger$  & (3m)$^\ddagger$  & 5.70$^\ddagger$ &  0.03\%$^\ddagger$  & (14m)$^\ddagger$  & 7.79$^\ddagger$  & 0.38\%$^\ddagger$  & (44m)$^\ddagger$\\ 
& POMO  \cite{kwon2020pomo} & 3.83$^\ddagger$  &  0.04\%$^\ddagger$  & (1s)$^\ddagger$  & 5.70$^\ddagger$ &  0.21\%$^\ddagger$  & (2s)$^\ddagger$  & 7.80  & 0.46\%  & {(12s)}\\ 
& POMO$\times$8 augment  \cite{kwon2020pomo} & 3.83$^\ddagger$  &  \textbf{0.00\%}$^\ddagger$  & (3s)$^\ddagger$  & 5.69$^\ddagger$ &  0.03\%$^\ddagger$  & (16s)$^\ddagger$  & 7.78  & {0.19\%}  & {(2m)}\\ 
\cmidrule(lr){2-11}
& CVAE-Opt-DE  \cite{hottunglearning}  & \multicolumn{1}{c}{-} & {$\approx$0.00\%}  & {($	\approx$1d)} &  \multicolumn{1}{c}{-} & {$\approx$0.02\%} & {($	\approx$2d)} & \multicolumn{1}{c}{-} & {$\approx$0.34\%} & {($	\approx$6d)} \\ 
& DPDP \cite{kool2021deep} (100k) & -  & -  & -  & -  &-& -  & 7.77$^\ddagger$ & 0.00\%$^\ddagger$ & (3h)$^\ddagger$ \\
 \midrule
 \midrule
\multirow{17}{*}{\rotatebox{90}{CVRP}} 
& {LKH}~\cite{lkh3} & 6.14  &  0.00\%  & {(18h)}  & 10.38 &  0.00\%  & {(3d)}  &{15.65}  & 0.00\%  & {(6d)} \\
& OR-Tools~\cite{ortools} & 6.46$^\ddagger$  &  5.68\%$^\ddagger$  & (2m)$^\ddagger$  & 11.27$^\ddagger$ &  8.61\%$^\ddagger$  & (13m)$^\ddagger$  &17.12$^\ddagger$  & 9.54\%$^\ddagger$  & (46m)$^\ddagger$ \\
\cmidrule(lr){2-11}
& NeuRewriter \cite{chen2019learning}  & 6.15$^\ddagger$ & \multicolumn{1}{c}{-}  & {(22m)$^\ddagger$} & 10.51$^\ddagger$ & \multicolumn{1}{c}{-}   & {(35m)$^\ddagger$}  & 16.10$^\ddagger$  & \multicolumn{1}{c}{-}  & {(1h)$^\ddagger$}  \\ 
& {NLNS~\cite{hottung2019neural}~(T=1k)} & {6.19} & {0.85\%}&{(9m)$^\#$}&{10.56}&{1.82\%}&{(14m)$^\#$}&{16.09}&{2.84\%}&{(31m)$^\#$} \\ 
& {NLNS~\cite{hottung2019neural}~(T=10k)} & {6.17} & {0.60\%}&{(1.5h)$^\#$}&{10.49}&{1.12\%}&{(2.5h)$^\#$}&{15.88}&{1.49\%}&{(5h)$^\#$} \\ 
& {\citet{wu2021learning} (T=1k)} & {6.16}$^\ddagger$  &  {0.90\%}$^\ddagger$  & {(23m)$^\ddagger$}  & {10.68} &  {2.89\%}  & {(50m)}  & {16.39}  & {4.76\%}  & {(1h)} \\ 
& {\citet{wu2021learning} (T=5k)} & 6.12$^\ddagger$  &  0.39\%$^\ddagger$  & (2h)$^\ddagger$  & {10.54} &  {1.63\%}  & {(4h)}  & {16.17}  & {3.31\%}  & {(5h)}\\ 
& DACT (T=1k) & {6.14}  & {0.04\%}  & {(4m)$[$1m$]$} & {10.55}  & {1.73\%}  & {(16m)$[$3m$]$}  & {16.20}  & {3.52\%}  & {(41m)$[$5m$]$}  \\
& DACT (T=5k) & 6.13  & {-0.07\%} & {(20m)$[$6m$]$} & {10.46}  & {0.80\%}  & {(1.5h)$[$13m$]$}  & {15.95} & {1.92\%}  & {(3.5h)$[$26m$]$}  \\ 
& DACT (T=10k) & 6.13 & {-0.07\%}  & {(40m)$[$12m$]$}  & {10.43}  & {0.58\%}  & {(2.5h)$[$25m$]$} & {15.88}  & {1.51\%}  & {(7h)$[$52m$]$}  \\ 
& DACT$\times$6 augment &6.13 & \textbf{-0.08\%}  & {(4h)$[$1h$]$}  & {10.38}  & {\textbf{0.08\%}}  &  {(16h)$[$2.5h$]$}  & {15.74} & {\textbf{0.57\%}}  & {(41h)$[$5h$]$}\\ 
\cmidrule(lr){2-11}
& AM-sampling \cite{kool2018attention} & 6.25$^\ddagger$  &  1.87\%$^\ddagger$ & (6m)$^\ddagger$  & 10.62$^\ddagger$ &  2.40\%$^\ddagger$  & (28m)$^\ddagger$  & 16.23$^\ddagger$ & 3.72\%$^\ddagger$  & (2h$^\ddagger$)\\ 
& MDAM-BS \cite{xin2020multi} & 6.14$^\ddagger$  &  0.18\%$^\ddagger$  & (5m)$^\ddagger$  & 10.48$^\ddagger$ &  0.98\%$^\ddagger$  & (15m)$^\ddagger$  & 15.99$^\ddagger$ & 2.23\%$^\ddagger$  & (1h)$^\ddagger$\\ 
& POMO \cite{kwon2020pomo} & 6.17$^\ddagger$  &  0.82\%$^\ddagger$  & (1s)$^\ddagger$ & 10.49$^\ddagger$ &  1.14\%$^\ddagger$  & (4s)$^\ddagger$  & {15.84}  & {1.21\%}  & {(21s)}\\ 
& POMO$\times$8 augment \cite{kwon2020pomo} & 6.14$^\ddagger$  &  0.21\%$^\ddagger$  & (5s)$^\ddagger$  & 10.42$^\ddagger$ &  0.45\%$^\ddagger$  & (26s)$^\ddagger$  & {15.75}  & {0.69\%}  & (2m)\\ 
\cmidrule(lr){2-11}
& CVAE-Opt-DE \cite{hottunglearning}   & {$\approx$6.14} &  \multicolumn{1}{c}{-}  & {($\approx$2d)} & {$\approx$10.40} &  \multicolumn{1}{c}{-} & {($\approx$5d)} & {$\approx$15.75}  & \multicolumn{1}{c}{-}  & {($\approx$11d)} \\ 
& DPDP \cite{kool2021deep} (100k)& -  & -  & - & -  & -  & - & 15.69$^\ddagger$ & 0.31\%$^\ddagger$ & (6h)$^\ddagger$ \\
\bottomrule
\end{tabular}
\begin{tablenotes}
\item[$*$] {We report the total time using one 2080 TI GPU or one CPU in (·), and the total time using multiple GPUs (4 for TSP, 8 for CVRP) in [·].}
\item[$\ddagger$] {The results (i.e., obj. values, gaps, and time) are adopted from its original paper or from \citet{wu2021learning}.}
\item[$\#$] {The run time of NLNS is based on one GPU and ten CPUs (in parallel) following its default setting.}
\end{tablenotes}
\end{threeparttable}%
}
\vspace{-8pt}
\end{table}

We evaluate our DACT model on two representative routing problems, i.e., TSP and CVRP~\cite{kool2018attention,wu2021learning, kwon2020pomo}. For each problem, we abide by existing conventions to randomly generate instances on the fly for three sizes, i.e., $N=$ 20, 50 and 100. 
Initial experiments with three operators including \emph{2-opt}, \emph{swap} and \emph{insert} show that \emph{2-opt} performs best for both TSP and CVRP (with \emph{insert} better than \emph{swap}), hence we report results of our method based on \emph{2-opt}. Following~\cite{wu2021learning,hottung2019neural,chen2019learning} we use randomly generated initial solutions for training and the solutions generated by the greedy algorithm for inference. Since each problem has its own constraints and node features, we adjust the input, feasibility masks, and problem-dependent hyperparameters for each problem, the details of which are provided in Appendix \ref{app:problem-formulation} and \ref{app:exp}. The DACT models and baselines are tested on a server equipped with RTX 2080 TI GPU cards and Intel Xeon E5-2680 CPU @ 2.40GHz. Our code in PyTorch are available here\footnote{\href{https://github.com/yining043/VRP-DACT}{https://github.com/yining043/VRP-DACT}}.

\subsection{Comparison studies}
In Table~\ref{tab:main}, we compare our DACT with, (1) learning based \emph{improvement} methods, including \citet{wu2021learning},  Neural-2-Opt~\cite{d2020learning} (TSP only), NeuRewriter~\cite{chen2019learning} (CVRP only), NLNS~\cite{hottung2019neural} (CVRP only), (2) learning based \emph{construction} methods, including AM-sampling~\cite{kool2018attention}, GCN-BS \cite{joshi2019efficient} (TSP only),  MDAM-BS~\cite{xin2020multi}, POMO~\cite{kwon2020pomo}, (3) conventional optimization algorithms equipped with learning based component(s), including DPDP~\cite{kool2021deep}, CVAE-Opt-DE \cite{hottunglearning}, and (4) strong conventional solvers including Concorde~\cite{concorde}, LKH~\cite{lkh2,lkh3}, and OR-Tools~\cite{ortools}. We do not include L2I~\cite{lu2019learning} as a baseline since it requires a prohibitively longer inference time than others\footnote{L2I needs 167 days for 10,000 CVRP100 instances, as estimated from 24min/instance in its original paper.}. All results are averaged over 10,000 randomly generated instances except for CVAE-Opt-DE (with $\approx$) which only infers 1,000 instances, 
and we report the metrics of objective values, (optimality) gaps and run time. We test strong baselines POMO (TSP100, CVRP100), Neural-2-Opt, Wu et al. (except for size 20), and NLNS on our machine based on the pre-trained models. Other results (with $\ddagger$) are adopted from their original papers or from~\citet{wu2021learning}.
For TSP, Concorde is adopted to get the optimal solutions; 
CVRP is harder to be solved optimally, and the gaps are based on objective values of LKH3. 
We report the total time using one GPU (for neural solvers) or one CPU (for conventional solvers) in~(·), and the total time using multiple GPUs (for DACT) in [·]; for NLNS (with $\#$), we use one GPU and ten CPUs (in parallel) following its default setting.
Note that the run time (for CPU solvers, and those with $\ddagger$ and $\#$) may not be directly comparable with ours and hard to compare due to various factors (e.g., GPU v.s. CPU, Python v.s. C, different devices) and thus we focus more on gaps.

Pertaining to TSP, our DACT with inference steps 1,000 (T=1k) and 5,000 (T=5k) significantly outperforms the traditional solver OR-Tools and the baseline \citet{wu2021learning} which directly adopted the original Transformer encoder. The DACT (T=5k) also deliveries lower gaps than construction methods including AM-sampling and GCN-BS on TSP100. With larger steps T=10k, our DACT further boosts the solution qualities and outperforms other construction methods including MDAM-BS (with beam search), and POMO (the current state-of-the-art). Besides, our DACT consistently outperforms improvement method Neural-2-Opt with shorter or similar run time. To further reduce the gaps, we also leverage the data augmentation technique in POMO (which considers flipping node coordinates without changing the optimal solution) to solve same instances multiple times in different ways. Although the inference time increases (we run data augmentation in serial), our DACT with 4 augments not only outstrips POMO with 8 augments but also achieves the lowest objective values and gaps among all purely learning based models.
In particular, we achieve the gap of 0.09$\%$ on TSP100, which is superior to most of the recent neural solvers.
Pertaining to CVRP, our DACT with T=5k achieves better objective values than that of NeuRewriter. It also consistently outperforms the important baseline \citet{wu2021learning} in all cases. Compared with the NLNS solver that learned to control the large neighborhood search heuristics, the gaps achieved by DACT (T=10k) are smaller on CVRP20 and CVRP50, and highly comparable on CVRP100.
With T=10k and 6 augments, our DACT exhibits even better performance than the highly specialized heuristic solver LKH on CVRP20 and delivers the smallest gap of 0.57\% on CVRP100 against other neural solvers including POMO with 8 augments.
Compared with hybrid solver CVAE-Opt-DE, despite that it is averaged over 1,000 instances and integrated with search algorithm like differential evolution, our objective values are still lower on both TSP and CVRP.
Nevertheless, our DACT is inferior to a more advanced hybrid solver DPDP which leverages learnt heat-maps and traditional dynamic programming to search solutions.

In terms of the inference time, our DACT is highly competitive against most of the neural solvers except for POMO which learns a \emph{construction} model by sampling diverse trajectories. However, when it comes to the generalization performance on benchmark datasets, i.e., TSPLIB~\cite{reinelt1991tsplib} and CVRPLIB~\cite{uchoa2017new} in Table \ref{tab:generalization}(a), DACT produces significantly lower average gaps than the POMO with 8 augments, which indicates that our DACT is more advantageous in practice despite its longer inference time. On the other hand, it is possible to adopt a similar diverse rollout strategy for DACT to find better solutions earlier, or explore other model compression techniques such as the knowledge distillation~\cite{hinton2015distilling} to learn a lighter DACT model for faster inference. Since our focus is to ameliorate Transformer for neural improvement solvers, we will investigate these possibilities in the future.

\begin{table}
\caption{Generalization performance. (a) DACT v.s. baselines on benchmark datasets (up to 200 customers, see Appendix \ref{app:Benchmark} for detailed results and discussion); (b) PE v.s. CPE on different sizes.}
\vspace{6pt}
\hspace{\fill}
\label{tab:generalization}
\begin{subtable}[t]{0.34\textwidth}
\centering
\resizebox{0.93\textwidth}{!}{
\begin{tabular}{@{}lcc@{}}
    \toprule
    Method  & {TSPLIB} & CVRPLIB \\ \midrule
    OR-Tools \cite{ortools} & 3.34\% & 8.06\% \\
    POMO \cite{kwon2020pomo} & 10.06\% & 6.10\% \\
    \midrule
    \citet{wu2021learning} & 4.17\% & 5.20\% \\
    {DACT-Avg} & {\textbf{1.90\%}} & {\textbf{3.85\%}} \\
     {DACT-Best} & {\textbf{1.01\%}} & {\textbf{2.97\%}} \\
    \bottomrule
\end{tabular}
}
\caption{}
\end{subtable}%
\hspace{\fill}%
\begin{subtable}[t]{0.45\textwidth}
\centering
\resizebox{\textwidth}{!}{
\begin{tabular}{@{}lcccc@{}}
            \toprule
            \multicolumn{1}{c}{\multirow{2}{*}{Method}} & \multicolumn{2}{c}{N=20} & \multicolumn{2}{c}{N=100} \\ 
             & Obj. & Gap & Obj. & Gap \\
             \midrule
            DACT-PE (T=5k) & 3.84 & 0.21\% & 8.38 & 7.93\% \\
            DACT-CPE (T=5k) & \textbf{3.83} & \textbf{0.10\%} & \textbf{7.99} & \textbf{2.98\%} \\
            \midrule
            \citet{wu2021learning} (T=5k) & 3.91 & 2.14\% & 9.03 & 16.37\% \\
            OR-Tools~\cite{ortools}  & 3.83 & 0.00\% & 8.06 & 3.87\% \\ \bottomrule
            \end{tabular}
}
\caption{}
\label{subtable1}
\end{subtable}%
\hspace{\fill}
\end{table}

\begin{table}
\vspace{-10pt}
\centering
\caption{Dual v.s. single aspect representation}
\label{tab:abalation}
\vspace{6pt}
\resizebox{0.5\textwidth}{!}{
\begin{tabular}{@{}rlccc@{}}
    \toprule
    Steps & Method & $\#$ Params  & N=50 & N=100 \\ \midrule
    \multirow{2}{*}{T=1k} & SA-T & 0.37M & 0.35\% {(2m)} & 3.49\% {(4m)}
     \\
     & DACT & 0.29M & \textbf{0.14\%} {(2m)} & \textbf{1.62\%} {(5m)}
     \\ \midrule
    \multirow{2}{*}{T=5k} & SA-T & 0.37M & 0.05\% {(7m)} & 1.55\% {(22m)} \\
    & DACT & 0.29M  & \textbf{0.02\%} {(8m)} & \textbf{0.61\%} {(25m)} \\ \bottomrule \\
\end{tabular}
}
\vspace{-10pt}
\end{table}

\subsection{Ablation studies}
\label{sec:ablation}
\paragraph{Dual-aspect representation.}
In Table \ref{tab:abalation}, we evaluate the effectiveness of our dual-aspect representation against the single-aspect one (SA-T) on TSP50 and TSP100, where SA-T mainly follows the Transformer in~\citet{wu2021learning} but equipped with the CPE, multi-head attentions and CL strategy for fair comparison. We observe that our DACT with fewer parameters consistently outperforms SA-T, which verifies the effectiveness of the dual-aspect representation.

\begin{figure*}
\centering 
     \subfloat[]{\includegraphics[width = 0.35\textwidth]{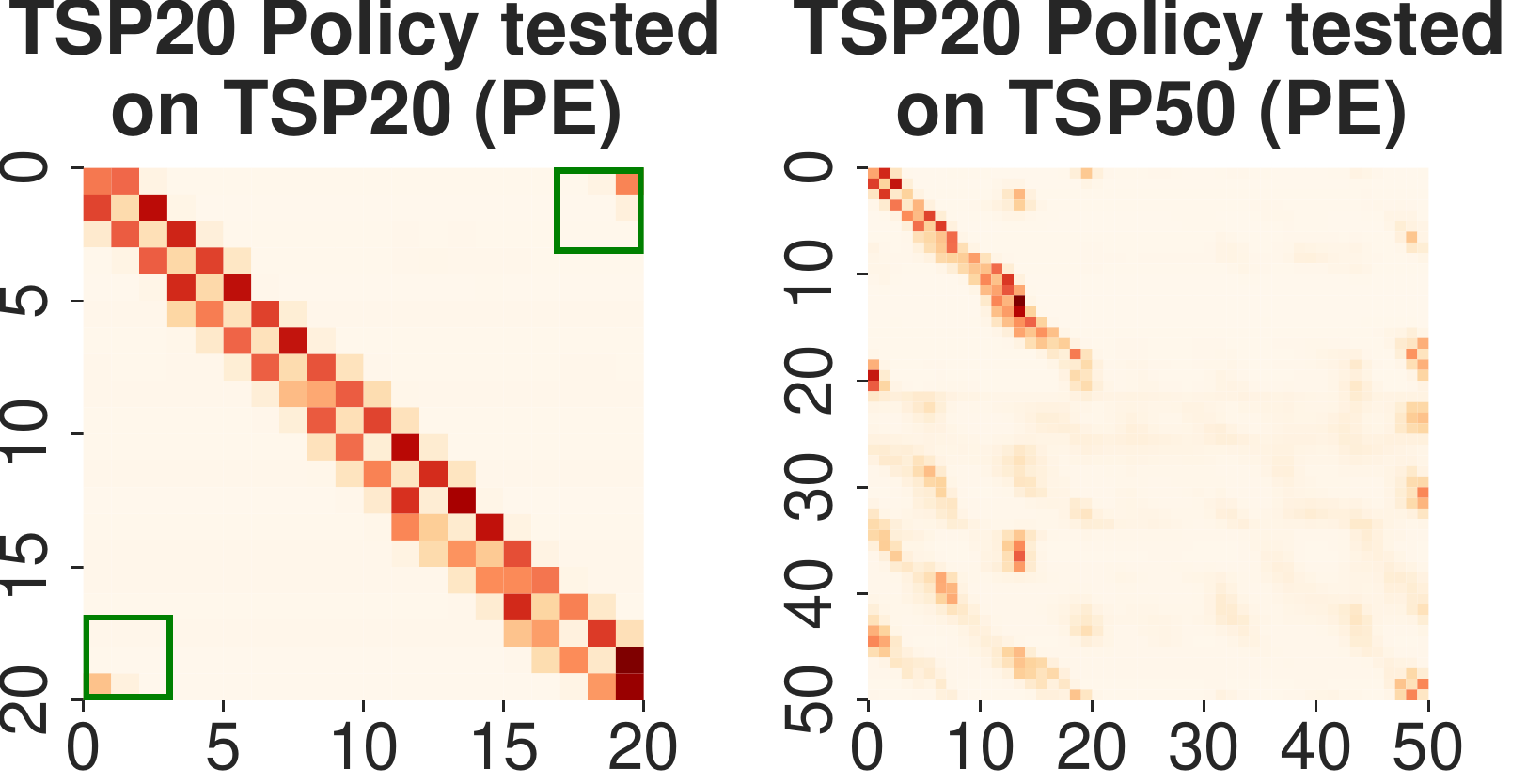}
    }
    \hspace{15pt}
     \subfloat[]{\includegraphics[width = 0.35\textwidth]{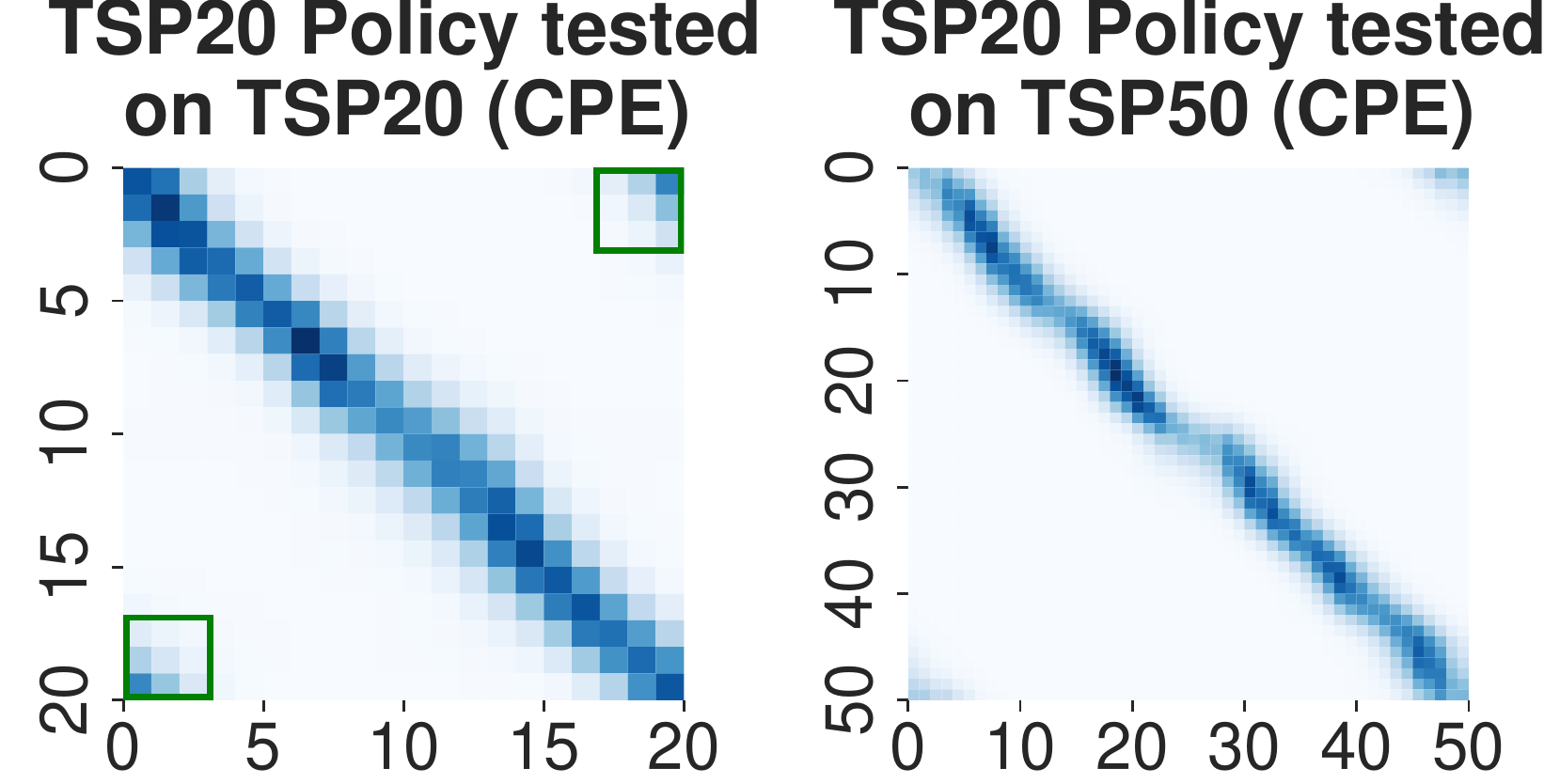}
    } 
	\caption{Visualization of the (multi-head) attention scores for the encoder when a trained model is used to solve instances with a larger size. (a) using PE method; (b) using CPE method (ours).}
	\label{fig:ab2}
\end{figure*}

\begin{wrapfigure}{r}{2.4in}
\centering 
	\includegraphics[width=1.85in]{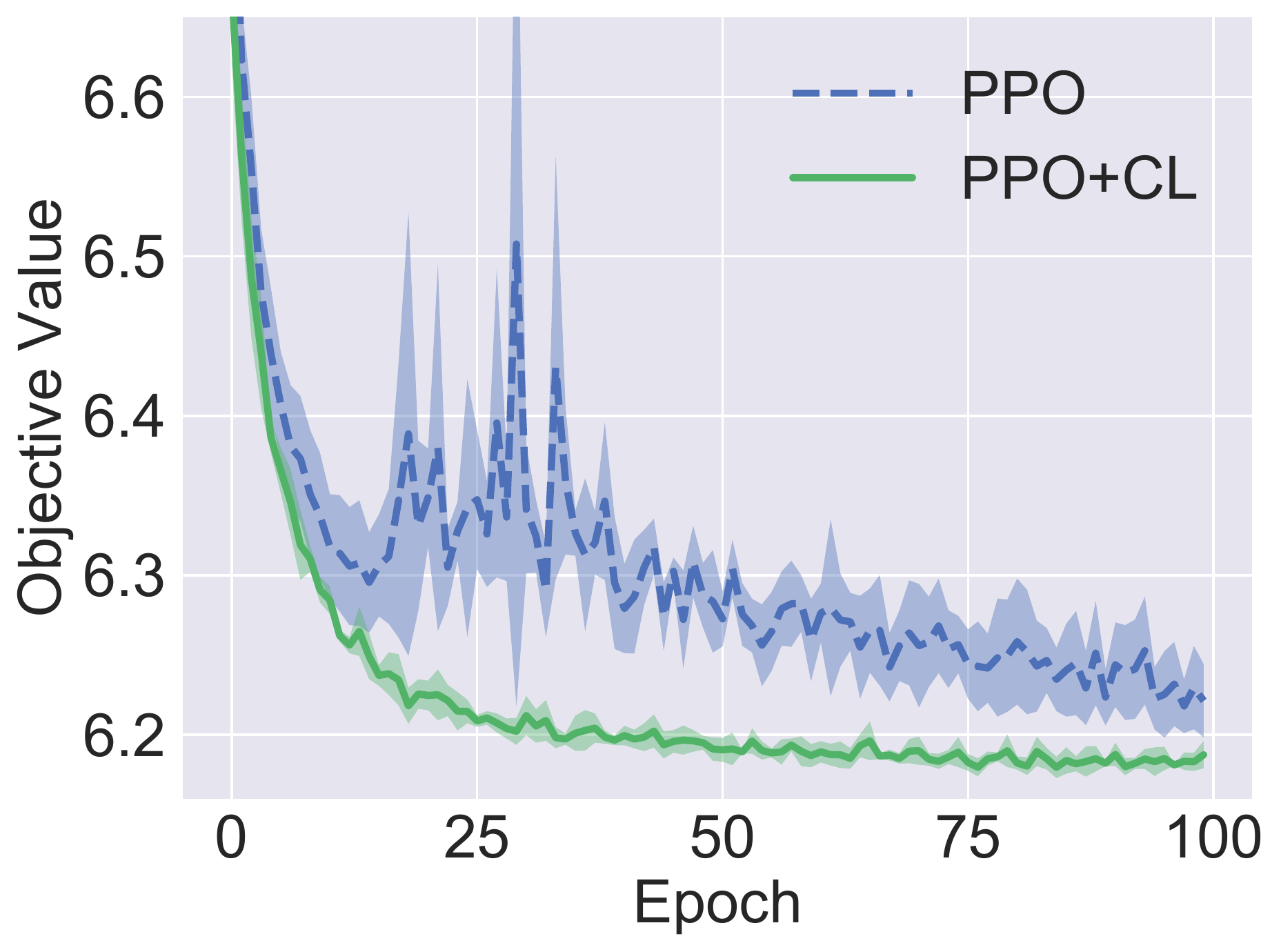} 
	\caption{{Training curves of PPO with and without CL on CVRP20 (random seeds 1-5).}} 
	\label{fig:CL}
\end{wrapfigure}
\paragraph{Cyclic positional encoding.} Here we show that CPE significantly improves the generalization performance across different problem sizes. In Table \ref{tab:generalization}(b), we record the results of our DACT with PE and CPE, and \citet{wu2021learning}, when the model trained on TSP50 is directly used to solve instances from TSP20 and TSP100 with T=5k. We see that even with PE, our DACT outperforms \citet{wu2021learning}. Further equipped with CPE, DACT outstrips DACT-PE and OR-Tools on TSP100. We continue to compare the two DACT variants by visualizing their attention scores. As depicted in Figure~\ref{fig:ab2}(a), although the absolute PE is designed for linear sequences, it did attempt to capture the circularity of VRP solutions (as highlighted in the green boxes) after training.
However, the ability to perceive such properties significantly drops when generalizing over different problem size, which instead engenders random attention scores when generalizing to larger size (see right side of Figure~\ref{fig:ab2}(a)). In contrast, our DACT with CPE is able to capture the circularity as depicted in Figure~\ref{fig:ab2}(b), which verifies the effectiveness of CPE in representing cyclic sequences (i.e., VRP solutions).

\paragraph{Curriculum learning (CL) strategy.}
In Figure \ref{fig:CL}, we plot training curves of PPO algorithm with and without our CL strategy for solving CVRP, where the results are averaged over 5 independent runs with 90\% confidence intervals. It shows that our CL strategy significantly improves the sample efficiency while reducing the variance of training, which aligns with our analysis in Section \ref{ppo-cl}.

\section{Conclusions and future work}
\label{sec:conclusion}
In this paper, we present a novel DACT model for routing problems. It learns separate groups of embeddings for the node and positional features, and is equipped with cyclic positional encoding (CPE) to capture the circularity and symmetry of VRP solutions. A curriculum learning (CL) strategy is also exploited to improve the RL training efficiency. Extensive experiments on both synthetic and benchmark datasets justified the effectiveness of DACT in terms of both inference and generalization. A potential limitation is that DACT is more useful for learning improvement models at present. In the future, we will investigate how to extend DACT to construction models, and how to speed up the DACT through diverse rollouts or model compression techniques. It is also interesting to apply the proposed CPE to develop Transformer based model for other tasks where the cyclic property is also important, e.g., encoding circular DNA/RNA structures in computational biology \cite{yu2017circular,liu2019biogenesis}.

\begin{ack}
This work was supported in part by the National Natural
Science Foundation of China under Grant 61803104 and Grant 62102228,
in part by the Young Scholar Future Plan of Shandong University under
Grant 62420089964188, and in part by the A*STAR CyberPhysical Production System (CPPS) - Towards Contextual and
Intelligent Response Research Program, under the RIE2020
IAF-PP Grant A19C1a0018, and Model Factory@SIMTech.

\end{ack}


{\small
\bibliographystyle{unsrtnat}
\bibliography{main}
}


\newpage
\appendix
\setcounter{page}{1} 

\vbox{
\hrule height 4pt
\vskip 0.25in
\vskip -\parskip%
\centering
{\LARGE\bf Learning to Iteratively Solve Routing Problems with Dual-Aspect Collaborative Transformer (Appendix)}
\vskip 0.29in
\vskip -\parskip
\hrule height 1pt
}
  
\section{Issues in existing Transformer-based model for VRP}
\label{app:unified}
\begin{figure}[H]
\centering 
	\includegraphics[width=1\textwidth]{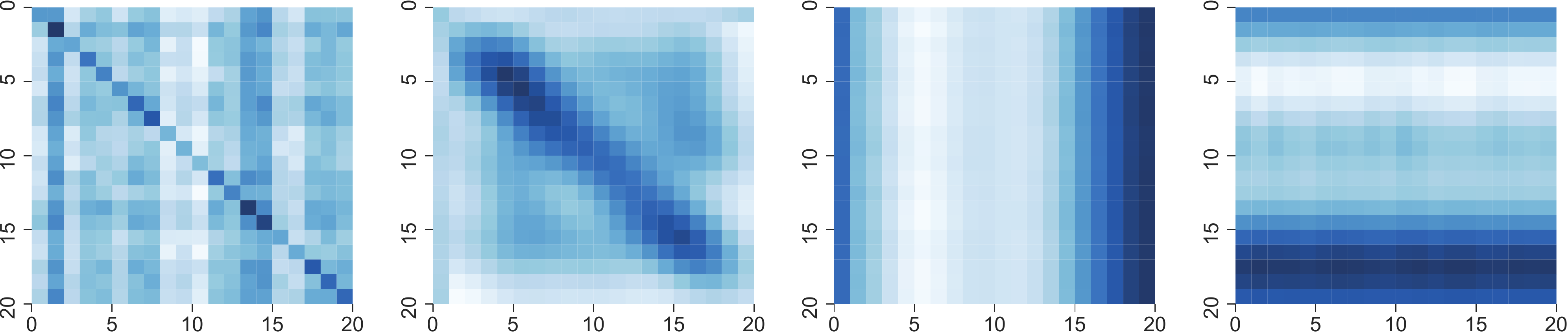} 
	\captionsetup{justification=justified}
	\caption{Visualizations of correlations on a trained model of \citet{wu2021learning}. From left to right: correlation for node-to-node, position-to-position, node-to-position, and
position-to-node, respectively.}
	\label{fig:correlation} 
\end{figure}

We investigate the issues of mixed correlations and noisy biases when the absolute PE method of Transformer is directly used to learn \emph{improvement} heuristics in \citet{wu2021learning}. By fusing the node feature embedding $h_i$ and the positional feature embedding $g_i$ through an addition operator, four attention query terms from input $i$ to $j$ exist during the self-attention of the encoder as follows, 
\begin{equation}
\label{eq:mixed4}
\begin{split}
    \alpha_{i,j}^{Abs} &=  \frac{1}{\sqrt{d_k}} \left( (h_i + g_i) W ^{Q} \right ) \left( (h_j + g_j)  W^{K} \right) ^T \\
    &= \frac{1}{\sqrt{d_k}} \left( h_i W ^{Q} \right ) \left( h_j  W^{K} \right) ^T 
    + \frac{1}{\sqrt{d_k}} \left( g_i W ^{Q} \right ) \left( g_j  W^{K} \right) ^T \\
    &+  \frac{1}{\sqrt{d_k}} \left( h_i W ^{Q} \right ) \left( g_j  W^{K} \right) ^T 
    +  \frac{1}{\sqrt{d_k}} \left( g_i W ^{Q} \right ) \left( h_j  W^{K} \right) ^T, 
\end{split}
\end{equation}
where we call them \emph{node-to-node}, \emph{position-to-position}, \emph{node-to-position}, and \emph{position-to-node}, respectively. Obviously, they all share the same projection matrices $W^{Q}$ and $W^{K}$, which might be unreasonable since they are used to represent correlations of different information~\cite{ke2020rethinking}.
Furthermore, the last two terms are essentially computing the mixed correlations across different information. Intuitively, queries from the location of a node (node feature) to the index of another node (positional feature) would be meaningless and vice versa. Such design may further bring noisy biases to routing problems. To verify this, we visualize the above four attention terms using a pre-trained model of \citet{wu2021learning} on a sampled batch of instances for TSP20. As shown in Figure~\ref{fig:correlation}, the last two correlations (node-to-position and position-to-node) seem to unreasonably present some random patterns across different node pairs, e.g., all nodes tend to have strong correlations with the ones appeared close to the end of the solution. This may yield biased attention, and thus affect the accuracy and the performance of the learned heuristics. In contrast, our DACT avoids such mixed correlations by learning feature embeddings for two aspects separately without fusing them into a unified representation.

\section{Details of CPE}
\label{app:CPE}
We let the angular frequencies $\omega_d=\frac{2\pi}{\lambda_d}$ decrease along the vector dimension by increasing the wavelength $\lambda_d$ according to Eq.~(\ref{eq:freq}).
Empirically, we fix the last half of the wavelength to $N$ (the largest value) to better preserve the desired cyclic and adjacency similarity properties. Regarding the generalization of CPE, we explored two strategies: 1) directly use the target size $N_\text{new}$ to generate the above frequencies, 2) reuse the frequencies generated based on $N_\text{old}$ in pre-trained model. We find that the former is often better when the difference between $N_\text{new}$ and $N_\text{old}$ is not too large, and conversely, the latter could be better. In this paper, we report the best performance under both strategies.
\begin{equation}
\label{eq:freq}
\lambda_d\!=\!\!\left\{\begin{matrix}
\frac{\!3\lfloor d/3 \rfloor + 1}{dim} (N\!\!-\!\!N^{\frac{1}{\left \lfloor dim / 2 \right \rfloor}})\!+\! N^{\frac{1}{\left \lfloor dim / 2 \right \rfloor}}\!, \!\!\!\!&\!\text{\!if \!}d\!<\!\lfloor\!\frac{dim}{2}\!\rfloor\\ 
N,\!&\!\text{otherwise}
\end{matrix}\right.
\end{equation}

\section{Training algorithm}
\label{app:algorithm}
\begin{algorithm}[H]
\caption{$n$-step PPO with curriculum learning strategy}
\label{ppo}
\begin{algorithmic}[1]
\REQUIRE initial policy network parameters $\theta$; initial value function parameters $\phi$; clipping threshold $\varepsilon$; initial learning rate $\eta_\theta$, $\eta_\phi$, learning rate decay  $\beta$.
\FOR{$epoch = 1$ to $E$}
\FOR{$b = 1$ to $B$}
\STATE Randomly generate a batch of training instances $\mathcal{D}_b$;
\STATE Initialize random solutions $\{\delta_{i}\}$ to $\mathcal{D}_b$;
\STATE CL: Improve $\{\delta_{i}\}$ to $\{\delta'_{i}\}$ by iterating $T_{init}=\frac{S(e) - S(0)}{S(E) -S(0)} \xi^{CL} E$ steps with the current policy network (DACT) $\pi_\theta$, where $S(epoch) = 1 / \left({1 + e^{-\kappa(epoch - E/2)}} \right)$;
\STATE Set initial state $s_0 = \{\delta'_i\}$, $t \gets 0$;
\WHILE{$t < T_{train}$}
\STATE Collect experience $\{(s_{t'},a_{t'},r_{t'})\}^{t + n}_{t' = t}$ by running policy $\pi_\theta$ for $n$ time steps where  $a_{t'} \sim \pi_\theta(a_{t'}|s_{t'})$;
\STATE Set $t \gets t + n$, $\pi_{old} \gets \pi_\theta$, $v_{old} \gets v_\phi$;
\FOR{$k = 1$ to $K$}
\STATE $\hat R_{t + 1} = v_\phi(s_{t + 1})$;
\FOR{$t' \in \{t, t - 1, ...,  t - n\}$}
\STATE $\hat R_{t'} \gets r_{t'} + \gamma \hat R_{t'+1}$;
\STATE $\hat A_{t'} \gets \hat R_{t'} - v_\phi(s_{t'})$;
\ENDFOR
\STATE Compute PPO-Clip objective $J_{PPO}(\theta)$ using Eq.~(\ref{eq:ppo}) and clipped value function loss $L_{BL}(\phi)$ using Eq.~(\ref{eq:BL});
\STATE $\theta \gets \theta + \eta_\theta \nabla J_{PPO}(\theta)$;
\STATE $\phi \gets \phi - \eta_\phi \nabla L_{BL}(\phi)$;
\ENDFOR
\ENDWHILE
\ENDFOR
\STATE $\eta_\theta \gets \beta \eta_\theta$, $\eta_\phi \gets \beta \eta_\phi$;
\ENDFOR
\end{algorithmic}
\end{algorithm}
As presented in Algorithm \ref{ppo}, our training algorithm for DACT is adapted from the proximal policy optimization (PPO) \cite{schulman2017proximal}, which is a prevailing RL algorithm. In particular, we follow the actor-critic variant of PPO which considers subtracting a baseline $v_\phi(s_t)$ (i.e., value function) in the objective function (line 14) to reduce the variance.
Our $v_\phi$ is similar to the one in \citet{wu2021learning} as follows, (1) it takes the concatenation of node and positional embeddings as input, and then enhances them by a normal multi-head attention layer (with 6 heads); and (2) the enhanced embeddings are passed through a mean-pooling layer (similar to the max-pooling layer in the DAC decoder) and then processed by a four-layer feed forward network (with 128 and 64 hidden units) to get the output value. 

We train $\pi_\theta$ and $v_\phi$ for $E$ epochs and $B$ batches per epoch. For each batch, we generate training instances $\mathcal{D}_b$ on the fly (line 3) and use the proposed \textit{curriculum learning} strategy to initialize the state (line 4 to 6). We exploit the $n$-step return estimation to attain a satisfactory trade-off between one step temporal difference (TD) method and Monte Carlo (MC) method~\cite{wu2021learning} (line 8 to 15). Afterwards, PPO performs $K$ epochs of updates on $\mathcal{D}_b$ with its objective clipped by a threshold $\varepsilon$ to penalize large policy variances that move the probability ratio $\frac{\pi_\theta(a_{t}|s_{t})}{\pi_{old}(a_{t}|s_{t}}$ away from 1 (as shown in Eq.~(\ref{eq:ppo})), 
\begin{equation}
\label{eq:ppo}
 \begin{split}
    J_{PPO}(\theta)\!=\!\frac{1}{|\mathcal{D}_b|n}\sum_{\mathcal{D}_b}\sum_{t' = t}^{t+n}min\left( \frac{\pi_\theta(a_{t'}|s_{t'})}{\pi_{old}(a_{t'}|s_{t'})}\hat A_{t'}, \right.
    \left. \!clip \left[ \frac{\pi_\theta(a_{t'}|s_{t'})}{\pi_{old}(a_{t'}|s_{t'})}, 1-\varepsilon, 1 + \varepsilon \right] \hat A_{t'}\right).
\end{split}   
\end{equation}

We also clip the estimated value $\hat v(s_t)$ around the previous one as shown in Eq.~(\ref{eq:clipp}) for better performance~\cite{engstrom2020implementation} and define the baseline loss in Eq.~(\ref{eq:BL}). The parameters of our two networks are updated by the Adam Optimizer (line 17,18) with a decaying learning rate (line 22),
\begin{equation}
\label{eq:clipp}
    v_\phi^{clip}(s_{t'})= clip\left[ v_\phi(s_{t'}), v_{old}(s_{t'})-\varepsilon, v_{old}(s_{t'})+\varepsilon \right],
\end{equation}
\begin{equation}
\label{eq:BL}
 \begin{split}
    L_{BL}(\phi) = \frac{1}{|\mathcal{D}_b|n}\sum_{\mathcal{D}_b}\sum_{{t'} = t}^{{t}+n}max\left( \left| v_\phi(s_{t'}) - \hat R_{t'}\right|, \right. 
    \left.\left| v_\phi^{clip}(s_{t'}) - \hat R_{t'}\right| \right)^2.
\end{split}   
\end{equation}

For our curriculum learning strategy, we adopt $\kappa\!=\!0.2$ and adjust the maximum CL step limit coefficient $\xi^{CL}$ according to the difficulty level and the problem sizes of the routing problems. Ideally, the selected $\xi^{CL}$ should satisfy the following conditions: (1) it is able to considerably boost the sample efficiency of training compared with the smaller ones, and (2) if a larger value is adopted, it may not be able to further bring a significant improvement. In practice, we recommend to determine the value by performing preliminary short training (around 10 epochs) with different $\xi^{CL}$.

\section{Problem-specific description}
\label{app:problem-formulation}
\subsection{Travelling salesman problem (TSP)}
\paragraph{Problem setup.}
A TSP instance considers to find the shortest loop that visits $N$ nodes exactly once, and finally returns to the original one. We follow \citet{kool2018attention} to generate the coordinates of $N$ nodes in the unit square $[0, 1] \times [0, 1]$ with a uniform distribution.

\paragraph{State feature representations.}
The node feature $x_i$ of node $i$ in the state is represented by its location. Let $c_i$ denote the 2-dim coordinates of node $i$. We define $x_i\!=\!c_i$.

\subsection{Capacitated vehicle routing problem (CVRP)}
\paragraph{Problem setup.}
On the basis of TSP, we add another node with index $0$ as the depot, and let the original $N$ nodes be customers. A CVRP instance considers to find the minimum total travel distance to serve all customers with multiple vehicles. Hence the solution to CVRP may consist of multiple sub-routes, each of which is a loop of nodes visited by one vehicle, i,e., departing from the depot, visiting the customers on this sub-route, and finally returning to the depot.
The constraints of CVRP include: 1) the total demands of a sub-route cannot exceed the vehicle capacity $Q$ , and 2) all customers must be visited exactly once. Similarly, we follow \citet{kool2018attention} to generate the coordinates of all nodes in the unit square $[0, 1] \times [0, 1]$ with the uniform distribution, and sample the demand of each customer uniformly from $\{1, 2, ..., 9\}$. The $Q$ is set to be 30, 40 and 50 for CVRP20, CVRP50 and CVRP100, respectively.

The length of CVRP solution might be larger than $N\!+\!1$ since the depot could be visited for multiple times. It may also vary even for the same instance, since different solutions may contain different numbers of sub-routes. For example, both $\delta_0\!=\!\{0,1,2,0,4,3,0\}$ and $\delta_1\!=\!\{0,1,2,3,4,0\}$ with different lengths could be feasible solutions to CVRP with 4 customers. Such varying lengths render it much hard for parallel batch training. We thus add multiple dummy depots to the end of initial solutions following~\citet{wu2021learning}, where the number of dummy depots could be also considered as the maximum number of available vehicles. In the aforementioned example, after adding dummy depots (we index the depots as (1),(2),(3)), the solution $\delta_1'\!=\!\{0^{(1)},1,2,3,4,0^{(2)},0^{(3)}\}$ will have the same length with $\delta_0$. In doing so, 1) it guarantees the same length of solutions for a instance batch; and 2) it allows the policy to automatically learn the number of sub-routes and their lengths in a solution. E.g., $\delta_1'\!=\!\{0^{(1)},1,2,3,4,0^{(2)},0^{(3)}\}$ at step $t$ could be changed to $\{0^{(1)},1,2,0^{(2)},4,3,0^{(3)}\}$ (equivalent to $\delta_0$) at step $t\!+\!1$ using the 2-opt operator given action $(3,0^{(2)})$. In our experiments, we empirically set 10 (dummy) depots for CVRP20, and 20 (dummy) depots for CVRP50 and CVRP100, respectively.

\paragraph{State feature representations.}
For CVRP, the node feature $x_i$ for node $i$ in the state is represented as a 7-dimensional vector \cite{wu2021learning}, which contains, 1) the 2-dim coordinates of node $i$, i.e., $c_i$; 2) the distance from node $i$ to its preceding neighbour; 3) the distance from node $i$ to its succeeding neighbour; 4) sum of demands of the corresponding route before node $i$; 5) the demand of node $i$;  and 6) sum of demands of the corresponding route after node $i$ (including node $i$). Due to the  circularity and symmetry of VRP solutions, we consider the succeeding neighbour of the last node in a solution to be the first node, and the preceding neighbour of the first node to be the last node.

\section{More discussion on the experiments}
\label{app:exp}

\subsection{Hyperparameters}

We set $\xi^{CL}$ as 0.25, 2, 10 for TSP 20, TSP50, and TSP100; 1, 4, 12.5 for CVRP20, CVRP50, and CVRP100, respectively. To avoid exploding gradients, we follow~\cite{kool2018attention,xin2020multi,li2021heterogeneous} to clip the gradient norm of each model parameter to be within 0.04, 0.2, and 0.45 for both TSP and CVRP of the three sizes, respectively. We set the reward discount factor $\gamma = 0.999$ for both problems. Regarding $T_r$, we use 250 for inference; 10 (TSPLIB) and 20 (CVRPLIB) for generalization on benchmark datasets.

\subsection{Training}
We train our model with $E\!=\!200$ epochs and $B\!=\!20$ batches per epoch with batch size 600 for TSP and CVRP. We use 512 for CVRP100 due to the limited GPU memory. Regarding the $n$-step PPO algorithm, we set $n\!=\!4$, $T_{train}\!=\!200$  for TSP, and $n\!=\!5$,  $T_{train}\!=\!250$ for CVRP. PPO performs $K\!=\!3$ mini-batch updates per batch with its objective function clipped by a threshold $\epsilon\!=\!0.1$. We adopt the Adam optimizer with a learning rate $\eta_\theta\!=\!10^{-4}$ for $\pi_\theta$ and $\eta_\phi\!=\!3\!\times\!10^{-5}$ for $v_\phi$, both of which are decayed with $\beta\!=\!0.985$ per epoch for convergence. We use pretrained models for TSP50, CVRP20, and CVRP50 to train TSP100, CVRP50, and CVRP100 for faster convergence, while for others the model is initialized randomly. We enable all state transitions of the MDP and the masking for feasibility of a batch to be performed in parallel on GPU for higher efficiency. Training time depends on the problem and varies with problem sizes. For example, the training time for CVRP sizes 20, 50, and 100 are around 4 days (1 GPU), 1 week (2 GPUs), and 2 weeks (4 GPUs), respectively.

\begin{figure}
     \centering
     \subfloat[]{\includegraphics[width = 0.4\textwidth]{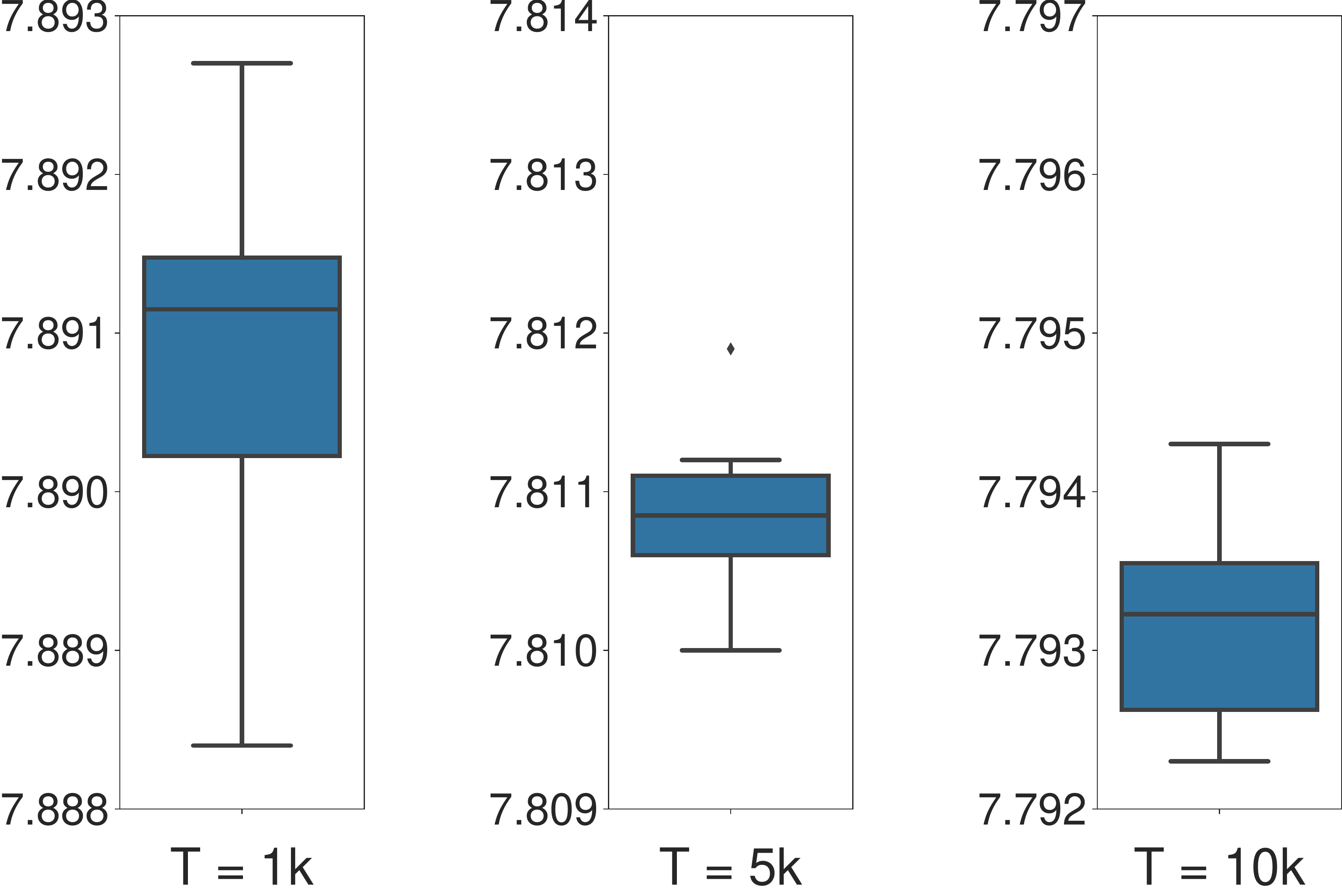}}
     \hspace{20pt}
     \subfloat[]{\includegraphics[width = 0.4\textwidth]{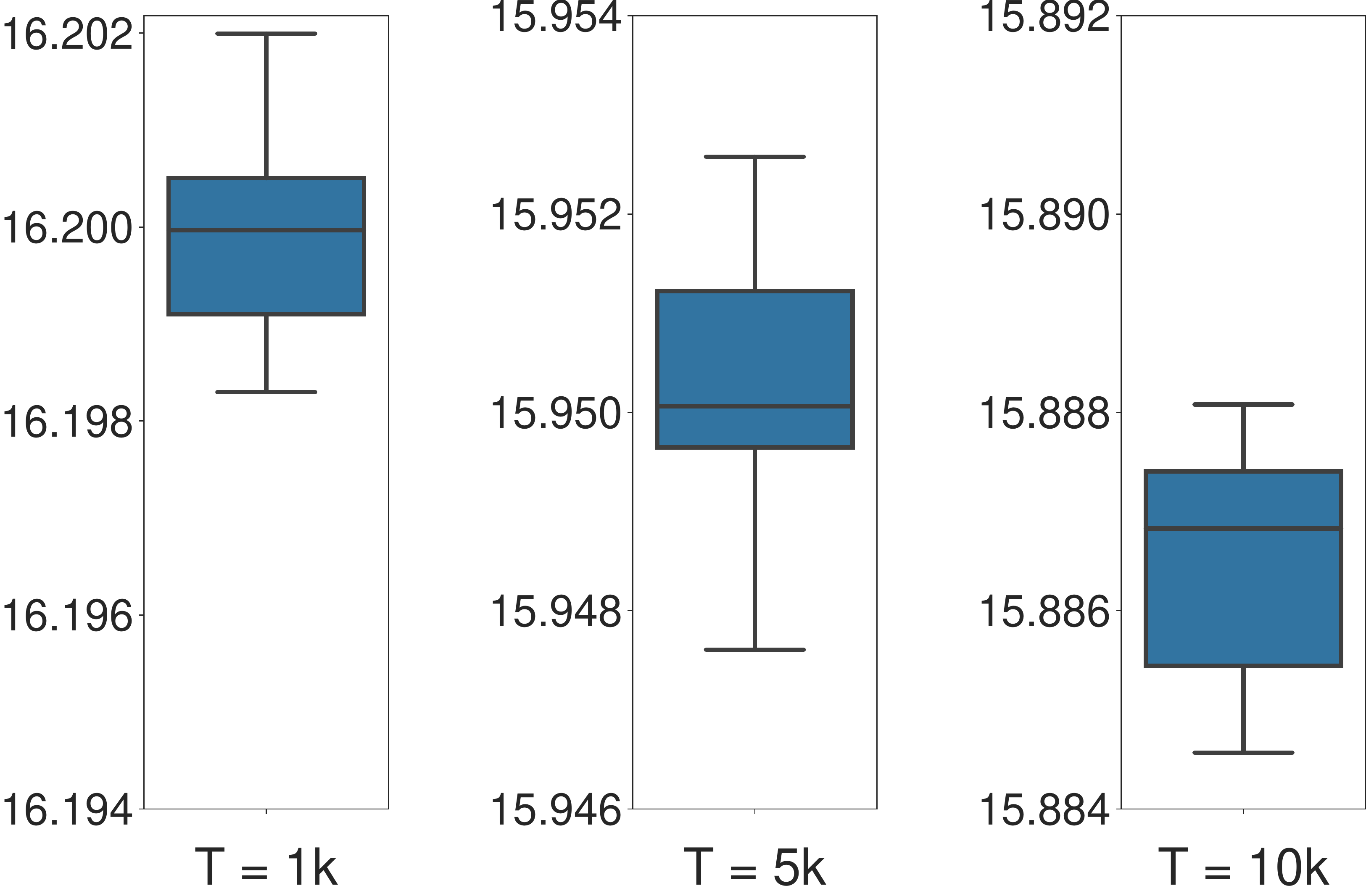}}
     \caption{Box plots of the objective values obtained by our DACT model (without augments) for 10 independent runs of 10,000 testing instances (with random seeds 1-10). (a) TSP100; (b) CVRP100.}
     \label{fig:errorbar}
\end{figure}

\begin{table*}[t]
\centering
\caption{Generalization of DACT (10 runs) v.s. baselines on TSPLIB benchmark dataset.}
\label{tab:tsplib}
\resizebox{1\textwidth}{!}{%
\begin{threeparttable}
\begin{tabular}{@{}c||ccc||cccc||cccc@{}}
\toprule
\midrule
\multirow{2}{*}{Instance} &
  \citet{wu2021learning} &
  \multicolumn{2}{c||}{DACT(T=3k)} &
  \multirow{2}{*}{OR-Tools} &
  AM-sampling &
  POMO &
  \citet{wu2021learning} &
  \multicolumn{2}{c}{DACT(T=10k)} &
  \multicolumn{2}{c}{DACT($\times$4 augment)} \\
         & (T=3k)  & Best   & Avg     &        & (N=10k)  & $\times$8 augment & (T=3k, M=1k) & Best   & Avg    & Best   & Avg    \\
\midrule
\midrule
eil51    & 2.82\%  & 0.23\% & 1.06\%  & 2.35\% & 2.11\%   & 0.00\%  & 1.17\% & 0.23\% & 0.87\% & 0.23\% & 0.26\% \\
berlin52 & 6.34\%  & 0.00\% & 0.94\%  & 5.34\% & 1.67\%   & 0.03\%  & 2.57\% & 0.00\% & 0.53\% & 0.00\% & 0.00\% \\
st70     & 4.59\%  & 0.00\% & 0.21\%  & 1.19\% & 2.22\%   & 0.30\%  & 0.89\% & 0.00\% & 0.13\% & 0.00\% & 0.00\% \\
eil76    & 6.88\%  & 0.74\% & 1.90\%  & 4.28\% & 3.35\%   & 1.49\%  & 4.65\% & 0.00\% & 1.39\% & 0.00\% & 0.72\% \\
pr76     & 1.40\%  & 0.03\% & 0.75\%  & 2.72\% & 2.84\%   & 19.97\% & 1.37\% & 0.00\% & 0.59\% & 0.00\% & 0.59\% \\
rat99    & 17.18\% & 1.98\% & 3.48\%  & 1.73\% & 9.50\%   & 7.51\%  & 8.51\% & 0.74\% & 2.58\% & 0.74\% & 1.29\% \\
KroA100  & 18.39\% & 0.43\% & 2.08\%  & 0.78\% & 79.49\%  & 4.45\%  & 2.08\% & 0.41\% & 1.13\% & 0.00\% & 0.21\% \\
KroB100  & 19.97\% & 0.26\% & 1.17\%  & 3.91\% & 9.30\%   & 5.83\%  & 5.78\% & 0.26\% & 0.67\% & 0.26\% & 0.38\% \\
KroC100  & 22.14\% & 0.39\% & 3.55\%  & 4.02\% & 8.04\%   & 6.55\%  & 3.17\% & 0.39\% & 2.56\% & 0.00\% & 0.77\% \\
KroD100  & 16.33\% & 0.57\% & 2.69\%  & 1.61\% & 10.02\%  & 8.74\%  & 5.00\% & 0.45\% & 2.23\% & 0.45\% & 0.64\% \\
KroE100  & 21.91\% & 1.10\% & 1.91\%  & 2.40\% & 3.10\%   & 5.97\%  & 3.29\% & 0.46\% & 1.14\% & 0.28\% & 0.60\% \\
rd100    & 0.06\%  & 0.00\% & 1.47\%  & 3.53\% & 1.93\%   & 0.00\%  & 0.06\% & 0.00\% & 0.99\% & 0.00\% & 0.59\% \\
eil101   & 4.61\%  & 1.91\% & 2.94\%  & 5.56\% & 3.97\%   & 2.07\%  & 4.61\% & 1.59\% & 2.23\% & 1.27\% & 1.75\% \\
lin105   & 26.53\% & 3.24\% & 7.23\%  & 3.09\% & 32.13\%  & 12.00\% & 2.48\% & 2.69\% & 6.37\% & 0.00\% & 0.89\% \\
pr107    & 19.76\% & 4.04\% & 4.97\%  & 1.74\% & 43.26\%  & 5.66\%  & 3.87\% & 2.87\% & 4.33\% & 2.87\% & 4.07\% \\
pr124    & 11.82\% & 0.66\% & 1.04\%  & 5.91\% & 4.41\%   & 0.29\%  & 2.97\% & 0.65\% & 0.75\% & 0.00\% & 0.30\% \\
bier127  & 20.65\% & 2.98\% & 4.91\%  & 3.76\% & 1.71\%   & 60.56\% & 3.48\% & 2.01\% & 3.39\% & 1.71\% & 2.43\% \\
ch130    & 16.53\% & 0.77\% & 3.64\%  & 2.85\% & 2.96\%   & 0.25\%  & 4.89\% & 0.41\% & 2.72\% & 0.41\% & 1.09\% \\
pr136    & 9.14\%  & 2.99\% & 6.71\%  & 5.62\% & 4.90\%   & 1.06\%  & 6.33\% & 1.45\% & 5.65\% & 0.58\% & 4.58\% \\
pr144    & 21.30\% & 2.30\% & 2.93\%  & 1.28\% & 8.77\%   & 0.80\%  & 1.40\% & 2.07\% & 2.63\% & 0.44\% & 1.16\% \\
ch150    & 21.26\% & 0.74\% & 1.86\%  & 3.08\% & 3.45\%   & 0.83\%  & 3.55\% & 0.58\% & 1.77\% & 0.87\% & 1.23\% \\
KroA150  & 17.80\% & 3.32\% & 6.12\%  & 4.03\% & 9.98\%   & 13.15\% & 4.51\% & 1.74\% & 4.21\% & 0.66\% & 5.15\% \\
KroB150  & 20.20\% & 2.84\% & 4.09\%  & 5.52\% & 9.87\%   & 11.72\% & 5.40\% & 1.64\% & 3.22\% & 1.64\% & 2.45\% \\
pr152    & 16.20\% & 1.84\% & 4.45\%  & 2.92\% & 13.47\%  & 4.11\%  & 2.17\% & 1.57\% & 3.10\% & 0.77\% & 2.23\% \\
u159     & 21.97\% & 2.50\% & 6.08\%  & 8.79\% & 7.38\%   & 2.19\%  & 7.67\% & 1.64\% & 5.13\% & 1.64\% & 2.69\% \\
rat195   & 25.40\% & 4.00\% & 4.87\%  & 2.84\% & 16.57\%  & 29.06\% & 9.90\% & 2.02\% & 3.92\% & 2.02\% & 3.28\% \\
d198     & 13.83\% & 7.05\% & 9.73\%  & 1.16\% & 331.58\% & 45.98\% & 4.99\% & 6.60\% & 8.62\% & 4.89\% & 6.31\% \\
KroA200  & 22.44\% & 4.46\% & 6.98\%  & 1.27\% & 15.64\%  & 20.00\% & 7.01\% & 3.05\% & 5.43\% & 2.33\% & 3.01\% \\
KroB200  & 23.69\% & 7.61\% & 10.90\% & 3.67\% & 18.54\%  & 21.06\% & 7.05\% & 5.46\% & 8.27\% & 5.23\% & 6.39\%\\
\midrule
\multicolumn{1}{l||}{Avg. Gap for [50,100)} & 6.53\% & 0.50\% & 1.39\% & 2.93\% & 3.61\% & 4.88\% & 3.19\% & 0.16\% & 1.01\% & 0.16\% & 0.48\%\\
\multicolumn{1}{l||}{Avg. Gap for [100,150)} & 9.69\% & 1.54\% & 3.37\% & 3.29\% & 15.29\% & 8.16\% & 3.53\% & 1.12\% & 2.63\% & 0.59\% & 1.39\% \\
\multicolumn{1}{l||}{Avg. Gap for [150,200]} & 12.76\% & 3.82\% & 6.12\% & 3.70\% & 47.39\% & 16.45\% & 5.81\% & 2.70\% & 4.85\% & 2.23\% & 3.64\%\\
\midrule
\multicolumn{1}{l||}{Avg. Gap for all instances}& 15.56\% & 2.03\% & 3.82\% & 3.34\% & 22.83\% & 10.06\% & 4.17\% & 1.41\% & 2.98\% & 1.01\% & 1.90\%  \\
\midrule
\bottomrule
\end{tabular}%
\end{threeparttable}
}
\end{table*}

\begin{table*}[]
\caption{Generalization of DACT (10 runs) v.s. baselines on CVRPLIB benchmark dataset.}
\label{tab:cvrplib}
\centering
\resizebox{\textwidth}{!}{%
\begin{tabular}{@{}ccc||ccc||cccc||cccc@{}}
\midrule
\midrule
\multirow{2}{*}{Instance} &
  \multicolumn{1}{l}{Depot} &
  \multicolumn{1}{l||}{Node} &
  \citet{wu2021learning} &
  \multicolumn{2}{c||}{DACT(T=5k)} &
  \multirow{2}{*}{OR-Tools} &
  AM-sampling &
  POMO &
  \citet{wu2021learning}  &
  \multicolumn{2}{c}{DACT(T=10k)} &
  \multicolumn{2}{c}{DACT($\times$6 augment)} \\
 &
  \multicolumn{1}{l}{Type} &
  \multicolumn{1}{l||}{Type} &
  (T=5k) &
  Best &
  Avg &
   &
  (N=10k) &
  $\times$8 augment &
  (T=5k, M=100) &
  Best &
  Avg &
  Best &
  Avg \\
  \midrule
  \midrule
X-n101-k25         & R & R  & 7.70\%  & 1.86\% & 3.49\%  & 6.57\%  & 32.95\% & 3.64\%  & 5.60\% & 1.68\% & 3.11\%  & 1.32\% & 1.86\% \\
X-n106-k14         & E & C  & 4.86\%  & 2.33\% & 3.00\%  & 3.72\%  & 6.78\%  & 1.85\%  & 2.83\% & 2.28\% & 2.77\%  & 1.72\% & 2.09\% \\
X-n110-k13         & C & R  & 6.39\%  & 0.41\% & 2.64\%  & 7.87\%  & 3.15\%  & 2.05\%  & 4.40\% & 0.37\% & 2.38\%  & 0.00\% & 1.10\% \\
X-n115-k10         & C & R  & 13.32\% & 1.23\% & 1.80\%  & 4.50\%  & 7.52\%  & 3.49\%  & 5.19\% & 0.08\% & 1.43\%  & 0.02\% & 0.68\% \\
X-n120-k6          & E & RC & 16.16\% & 3.52\% & 6.32\%  & 6.83\%  & 4.54\%  & 2.12\%  & 5.56\% & 3.49\% & 6.01\%  & 2.71\% & 4.00\% \\
X-n125-k30         & R & C  & 8.79\%  & 5.50\% & 6.71\%  & 5.63\%  & 35.16\% & 7.14\%  & 4.71\% & 4.76\% & 6.01\%  & 4.76\% & 5.51\% \\
X-n129-k18         & E & RC & 11.01\% & 4.06\% & 5.72\%  & 8.37\%  & 4.00\%  & 0.97\%  & 4.63\% & 3.98\% & 5.28\%  & 3.48\% & 3.91\% \\
X-n134-k13         & R & C  & 16.06\% & 4.85\% & 7.17\%  & 21.61\% & 20.13\% & 4.22\%  & 8.88\% & 3.77\% & 6.40\%  & 2.32\% & 3.28\% \\
X-n139-k10         & C & R  & 14.99\% & 1.82\% & 3.47\%  & 12.02\% & 4.30\%  & 2.28\%  & 4.90\% & 1.65\% & 3.05\%  & 0.53\% & 1.13\% \\
X-n143-k7          & E & R  & 20.20\% & 4.59\% & 5.86\%  & 11.27\% & 8.88\%  & 2.79\%  & 6.61\% & 3.92\% & 5.15\%  & 0.89\% & 3.37\% \\
X-n148-k46         & R & RC & 16.38\% & 4.09\% & 5.21\%  & 7.80\%  & 79.53\% & 19.88\% & 3.60\% & 3.61\% & 4.75\%  & 2.60\% & 4.24\% \\
X-n153-k22         & C & C  & 22.94\% & 6.10\% & 8.98\%  & 8.01\%  & 78.11\% & 12.16\% & 4.53\% & 4.63\% & 7.62\%  & 3.22\% & 4.26\% \\
X-n157-k13         & R & C  & 17.15\% & 3.24\% & 3.78\%  & 2.57\%  & 16.30\% & 2.79\%  & 3.60\% & 2.70\% & 3.36\%  & 2.09\% & 2.60\% \\
X-n162-k11         & C & RC & 19.16\% & 2.53\% & 3.98\%  & 6.31\%  & 6.37\%  & 4.77\%  & 5.26\% & 1.82\% & 2.68\%  & 1.10\% & 1.57\% \\
X-n167-k10         & E & R  & 18.52\% & 5.03\% & 6.36\%  & 9.34\%  & 8.41\%  & 4.05\%  & 8.27\% & 4.36\% & 5.75\%  & 4.05\% & 4.83\% \\
X-n172-k51         & C & RC & 12.06\% & 4.44\% & 6.50\%  & 10.74\% & 85.37\% & 21.99\% & 4.36\% & 3.66\% & 5.42\%  & 3.37\% & 3.78\% \\
X-n176-k26         & E & R  & 19.49\% & 8.53\% & 13.23\% & 8.99\%  & 20.39\% & 10.27\% & 6.16\% & 7.74\% & 10.91\% & 7.74\% & 9.13\% \\
X-n181-k23         & R & C  & 6.27\%  & 3.27\% & 3.84\%  & 2.94\%  & 6.45\%  & 2.08\%  & 2.08\% & 3.05\% & 3.65\%  & 1.89\% & 2.51\% \\
X-n186-k15         & R & R  & 17.71\% & 7.40\% & 9.15\%  & 7.75\%  & 6.01\%  & 2.15\%  & 7.65\% & 6.80\% & 8.01\%  & 5.01\% & 6.10\% \\
X-n190-k8          & E & C  & 18.64\% & 7.00\% & 8.69\%  & 6.53\%  & 46.61\% & 9.25\%  & 6.78\% & 6.36\% & 7.86\%  & 5.81\% & 6.42\% \\
X-n195-k51         & C & RC & 17.04\% & 6.78\% & 8.39\%  & 13.76\% & 79.26\% & 9.23\%  & 4.47\% & 6.60\% & 7.74\%  & 5.14\% & 6.25\% \\
X-n200-k36         & R & C  & 9.60\%  & 6.43\% & 7.26\%  & 4.15\%  & 26.25\% & 5.01\%  & 4.26\% & 6.30\% & 7.08\%  & 5.54\% & 6.07\% \\
\midrule
\midrule
\multicolumn{3}{l||}{Avg. Gap for [100,150)}   & 12.35\%   & 3.11\%        & 4.67\%         & 8.74\%                    & 18.81\%     & 4.58\%     & 5.17\%        & 2.69\%         & 4.21\%         & 1.85\%           & 2.83\%           \\
\multicolumn{3}{l||}{Avg. Gap for [150,200]}    & 16.24\%   & 5.52\%        & 7.29\%         & 7.37\%                    & 34.50\%     & 7.61\%     & 5.22\%        & 4.91\%         & 6.37\%         & 4.09\%           & 4.87\%           \\
\multicolumn{3}{l||}{Avg. Gap for all instances}     & 14.29\%   & 4.32\%        & 5.98\%         & 8.06\%                    & 26.66\%     & 6.10\%     & 5.20\%        & 3.80\%         & 5.29\%         & 2.97\%           & 3.85\%           \\ 
\midrule
\bottomrule
\end{tabular}
}
\end{table*}

\subsection{Stability analysis of our DACT}
We study the stability of our DACT model (without augments) during inference stage. Figure \ref{fig:errorbar} depicts the box plots of objective values on TSP100 and CVRP100 for 10 independent runs of 10,000 testing instances, where the minimum, lower quartile, mean, upper quartile, maximum, and possible outlets of results are depicted. In each sub-plot, we show the results for three step limit $T$, i.e., $T=$1k, 5k, and 10k as per the settings in Table \ref{tab:main}. For all cases, the range of box plots are within only 0.005 for both TSP and CVRP. These results show that our DACT has desirable stability for inference.

\subsection{Generalization on benchmark datasets}
\label{app:Benchmark}

We now evaluate the generalization performance of DACT by applying the trained models in Section~\ref{sec:exp} to solve instances from two well-known benchmark datasets, i.e., TSPLIB~\cite{reinelt1991tsplib} and CVRPLIB~\cite{uchoa2017new}, respectively. Note that these instances may follow completely different distributions from ours, such as clustered customer locations, corner depot location, etc. We report the results on instances with size between 50 and 200 for TSPLIB; and size between 100 and 200 for CVRPLIB.

As recorded in Table \ref{tab:tsplib} and \ref{tab:cvrplib}, we first compare our DACT with \citet{wu2021learning} in the first group of columns to verify the superior performance of DACT to the existing Transformer based improvement model. In the second group of columns, we report the performance of several strong baselines including, 1) OR-Tools~\cite{ortools}, 2) AM-sampling~\cite{kool2018attention}, 3) POMO $\times$8 augment~\cite{kwon2020pomo}, the state-of-the-art neural construction solver, and 4) the enhanced variant of \citet{wu2021learning}, which samples $M$ actions to produce multiple solutions at each step and retrieves the best one as the next state. We present the performance of our DACT with and without augments in the last group of columns.
For TSPLIB, we infer the first 5 instances (size $\!<\!99$) using DACT model trained on TSP50 and the remaining ones using model trained on TSP100. For CVRPLIB, we infer all instances using DACT model trained on CVRP100 since all the sizes are larger than 100. For AM-sampling and POMO, we use the trained models of our sizes which are provided by the authors. The results of \citet{wu2021learning} and OR-Tools are adapted from 
\citet{wu2021learning}.
The gaps are calculated based on the optimal solutions provided in the datasets. And we test our DACT for 10 runs and report the average and the best gaps. We also list the average gaps for instances in different problem size intervals, i.e., [50, 100), [100, 150) and [150, 200] for TSPLIB; and [100, 150) and [150, 200] for CVRPLIB.

\paragraph{TSPLIB}
Pertaining to TSPLIB in Table \ref{tab:tsplib}, our DACT (T=3k) significantly outperforms \citet{wu2021learning} (T=3k) for all instances expect for `rd100'. It also performs better than the two neural construction solvers AM-sampling (N=10k) and POMO$\times$8 augment for all three problem size intervals in terms of the average gaps, where the superiority to them becomes more obvious as the problem size increases. With larger steps (T=10k), our DACT continues improving the solution qualities and outstrips all the baselines including OR-Tools and~\citet{wu2021learning} (T=3k, M=1k), in terms of the overall average gap. Further empowered by 4 augments, our DACT consistently reduces the gaps and achieves the best performance on most instances with the lowest overall average gap.

\paragraph{CVRPLIB}
Pertaining to CVRPLIB in Table \ref{tab:cvrplib}, the depot and the nodes follow various distributions. Though trained on uniform distribution, our DACT (T=5k) outperforms \citet{wu2021learning} (T=5k), OR-Tools, AM-sampling (N=10k), and POMO$\times$8 augment in terms of gap on all instances. With T=10k and 6 augments, it further reduces the overall average gap to the lowest one.
This further verifies that our DACT generalizes well on real-world instances with various sizes and distributions.

\paragraph{Remarks}
It is worth noting that although POMO$\times 8$ augment previously achieved the state-of-the-art performance on synthetic instances according to \citet{kwon2020pomo}, it is still lacking in generalization on benchmark instances, whose underlying core model is AM (which showed the worst generalization performance in Table \ref{tab:main}). Meanwhile, we can also infer from the results that the neural improvement models including \citet{wu2021learning} and our DACT have much better generalization capability than the neural construction ones including AM-sampling and POMO. Given the advantages of our DACT model, it achieves the new state-of-the-art generalization performance among all existing Transformer based models on these benchmark instances from TSPLIB and CVRPLIB.

\subsection{Licenses for used assets}
We list the used existing assets in Table \ref{tab:license}. All of them are open-sourced assets for academic usage. Our code is released under the MIT open-source license.

\begin{table}[h]
\centering
\caption{List of used assets and the licenses.}
\label{tab:license}
\vspace{6pt}
\resizebox{0.6\textwidth}{!}{%
\begin{threeparttable}
\begin{tabular}{@{}llc@{}}
\toprule
Asset & Type & License \\ \midrule
OR-Tools \cite{ortools} & Code & Apache License, Version 2.0 \\
LKH, LKH3 \cite{lkh2,lkh3} & Code & Available for academic research use \\
AM \cite{kool2018attention} & Code & MIT License \\
NLNS \cite{hottung2019neural} & Code & GPL-3.0 License \\
Neural-2-Opt \cite{d2020learning} & Code & No License \\
\citet{wu2021learning} & Code & MIT License \\
POMO \cite{kwon2020pomo} & Code & MIT License \\
TSPLIB \cite{reinelt1991tsplib} & Dataset & Available for any non-commercial use \\
CVRPLIB \cite{uchoa2017new} & Dataset & Available for academic research use \\ \midrule
Python & Code & Python Software Foundation License \\
numpy & Code & BSD License \\
PyTorch & Code  & BSD License \\
tdqm & Code & MIT License \\
tensorboard & Code & Apache License 2.0\\
\bottomrule
\end{tabular}%
\end{threeparttable}
}
\end{table}

\end{document}